\documentclass[11pt]{article}

\usepackage{amsmath}
\usepackage{amssymb}
\usepackage{color}
\usepackage{graphicx}
\usepackage{url}

\usepackage{graphicx} 
\usepackage{url} 
\usepackage{multirow}
\usepackage{booktabs}
\usepackage{adjustbox}
\usepackage{array}
\usepackage{geometry}
\usepackage{makecell}
\usepackage{amsmath}
\usepackage{caption}
\usepackage{longtable}
\usepackage{enumitem}
\usepackage{cite}

\newtheorem{definition}{Definition}[section]

\newcommand{\cnp}{\textbf{NP}}

\begin{document}

\begin{center}
\Large{\textbf{Disjunctive and Conjunctive Normal Form Explanations}} \\ \smallskip
\Large{\textbf{of Clusters Using Auxiliary Information}} \\
\end{center}
\begin{center}
{Robert F. Downey}$^1$ \hspace*{0.1in}
{S. S. Ravi}$\,$$^{2}$ \\ \medskip
\date{March 2024}
\end{center}

\setcounter{footnote}{1}
\footnotetext[1]{
Department of Electrical and Computer Engineering,
University of Virginia, Charlottesville, VA 22904.
\textsf{Email:} \texttt{kuh4bd@virginia.edu}.
\medskip
}

\setcounter{footnote}{2}
\footnotetext[2]{
Biocomplexity Institute,
University of Virginia,
Charlottesville, VA 22904 ~and~
Department of Computer Science, University at Albany -- SUNY,
Albany, NY 12222.~
\textsf{Email:} \texttt{ssravi0@gmail.com}.
}


\medskip

\begin{center}
\textbf{Abstract}
\end{center}

\noindent
We consider generating post-hoc explanations of clusters generated
from various datasets using auxiliary information
which was not used by clustering algorithms.
Following terminology used in previous work,
we refer to the auxiliary information as tags.
Our focus is on two forms of explanations, namely disjunctive form
(where the explanation for a cluster consists of a set of tags)
and a two-clause conjunctive normal form (CNF) explanation 
(where the explanation consists of two sets of tags, combined
through the \textsc{And} operator).
We use integer linear programming (ILP) as well as heuristic methods
to generate these explanations.
We experiment with a variety of datasets and discuss the insights
obtained from our explanations.
We also present experimental results regarding the scalability 
of our explanation methods.

\bigskip


\section{Introduction} \label{sec:intro}

\subsection{Background, Motivation and Contributions}
\label{sse:motivation}

Tools based on Artificial Intelligence (AI) are currently being used in many business and societal applications such as obtaining loans from financial institutions, security checkpoints at airports and sentencing
decisions in courts 
(see e.g., \cite{adadi2018peeking,Zhang-Chen-2018,DBH-2018}).
When such applications use black box models, the danger
is that the results may be unfair to some sectors of
the population or cause harm.
As a consequence, the area of explainable AI (XAI) has become increasingly  important \cite{adadi2018peeking}.
A considerable amount of effort has been devoted to XAI for
supervised learning models (see e.g.,
\cite{xai-17-proc,xai-18-proc,Gunning-XAI-2017}).
The area of explainable AI for unsupervised learning
tasks such as clustering and classification has not
received as much attention \cite{DGR-NIPS-2018,Liu-etal-2022}.

\medskip

The focus of this paper is on explainability in the context
of clustering algorithms.
Specifically, we consider \emph{post~hoc} explanations of outputs
generated by clustering algorithms.
Our explanation method is based on the use of \emph{auxiliary
information} (or \textbf{tags}) associated with clustered data items.
The approach of using tags to explain clusters was proposed in
\cite{DGR-NIPS-2018} and was developed further in other
papers (see e.g., 
\cite{Sambaturu-etal-2020,Liu-etal-2022,davidson2020behavioral,bai2021towards}).
This approach constructs a short \textbf{descriptor}, 
which is a subset of tags, for each cluster.
A simple form of such a descriptor introduced
in~\cite{DGR-NIPS-2018} is a 
\textbf{disjunctive} descriptor,
where for each data item $x$ in a cluster, the corresponding
descriptor contains at least one tag that describes $x$.
Our work considers this form of descriptors as well as a more
general version called the 
\textbf{conjunctive normal form} (CNF) descriptor that
provides for two disjunctive descriptors.
While the CNF version requires more computation time, it has
the potential to provide better explanations  due to its expressive power.
Formal definitions concerning the descriptors are
presented in Section~\ref{sec:prelim}.

\medskip 

Our work constructs 
clusters of some public domain datasets
using the $k$-means
algorithm (see e.g., 
\cite{Han-etal-2012,Tan-etal-2019} for details regarding
this algorithm), 
generates disjunctive and CNF descriptors for the clusters and examines
the effectiveness of the descriptors.
Specifically, our contributions are as follows.

\begin{enumerate}
\item 
Since constructing optimal disjunctive descriptors is, in general,
computationally intractable \cite{DGR-NIPS-2018}, we use an
integer linear programming approach to develop descriptors
of minimum size.
As a baseline for comparison, we also use a greedy heuristic based on the \textbf{hitting set}
problem \cite{GJ-1979} for constructing disjunctive
descriptors.
In addition, we present a method for constructing
CNF descriptors.
To our knowledge,
the CNF form of descriptors has not been considered in the literature.

\item We construct descriptors for clusters from 
four public domain datasets and discuss the effectiveness
of the tag-based explanations. 
The sizes of these datasets
vary roughly from 170 to 30,000.

\item 
Examining the explanations for the datasets considered provides
interesting insights regarding the relative informativeness of
disjunctive and CNF explanations. The CNF form can aid in providing
additional tags when the disjunctive form provides little or
nondescript information. As an example, this added insight is
illustrated in Table~\ref{table:movies_hitting_sets} 
(Section~\ref{sse:movies}) for the
\textsf{Movies} dataset. Here, the CNF descriptor is formed by
combining the original disjunctive descriptor (generated by the
heuristic for hitting set) with another hitting set which is obtained
after excluding tags in the previous hitting set. This second hitting
set provides further information about the cluster beyond the one
or two tags in the disjunctive heuristic needed to cover every data
item in a cluster. More details are provided in 
Section~\ref{sse:movies}.

\item Using synthetic datasets, we examine the computation time
used to generate the descriptors to understand the scalability of
our methods. Our exact algorithm for generating a disjunctive
descriptor (based on an integer linear program) is slower on smaller
datasets, highlighting an even greater advantage of heuristic solvers
for these cases. For larger datasets, the CNF algorithm takes roughly
twice as long as the disjunctive solvers, as the disjunctive approach
is applied twice to generate the two clauses required for the CNF.
Additional details regarding this timing study appear in
Section~\ref{sse:scale}.
\end{enumerate}
More details regarding the contributions are provided 
in subsequent sections.


\subsection{Related Work}\label{sse:related}

As AI is beginning to be used extensively in many
applications, explainability of AI models has assumed
a lot of importance~\cite{adadi2018peeking,xai-17-proc,xai-18-proc,Miller-2018,Ming-2017,Zeng-etal-2018,Zhang-Chen-2018,DBH-2018}.
These references discuss methods used for both supervised
and unsupervised learning tasks.
Since our focus is on explaining outputs of clustering
algorithms, we will restrict our discussion to known
methods for that task. 
In the machine learning literature, explaining
clusters has been studied from two perspectives. 
The one-view approach of
\emph{conceptual clustering}~ \cite{frost2020exkmc,michalski1983learning,saisubramanian2020balancing} considers finding a clustering and its description simultaneously. Unlike our work, the
explanation is based on the features used for clustering.
So, these methods
require that the features used to perform clustering be
human interpretable.
Another approach, called \emph{exemplar-based 
clustering}~\cite{Davidson-etal-2024}, has also been
studied to generate a clustering and an explanation 
simultaneously. In this approach, the explanation for
a cluster consists of a small subset, called 
an \emph{exemplar}, of instances 
in that cluster such that every instance in the cluster
is close to at least one exemplar. This form of explanation is similar to the use of exemplars in cognitive psychology~\cite{Walsh-etal-2010}.

As mentioned earlier, the tag-based approach for explaining outputs of clustering algorithms was proposed in
\cite{DGR-NIPS-2018}.
That work established the complexity of obtaining 
descriptors of minimum length and provided an integer linear
programming approach for the problem.
Given the difficulty of constructing optimal descriptors
efficiently, Sambaturu et al.~\cite{Sambaturu-etal-2020}
present approximation algorithms for constructing descriptors of near-optimal length and established rigorous performance guarantees provided by 
the algorithms.
To decrease the length of cluster descriptors,
their approach allows the possibility that 
some data items in a
cluster may not be explained by the chosen descriptor.
Since our goal is to explain all the data items in
a cluster, our work does not use the algorithms
presented in \cite{Sambaturu-etal-2020}.
The tag based approach has been used in \cite{davidson2020behavioral} to explain the clusters
generated from the 2016 elections in France.
The approach has also been employed to explain results from the block model approach for community detection 
in networks \cite{bai2021towards}.
Liu et al.~\cite{Liu-etal-2022} consider a version of the
tag-based explanation problem with additional constraints and develop a quadratic programming formulation.
They develop algorithms which rely on specialized hardware to quickly generate descriptors for large problem instances.

In our work, the approach used to develop descriptors
relies on the close relationship between this problem
and the \textbf{hitting set problem}, which is a well
known \cnp-hard optimization problem~\cite{GJ-1979}.
This relationship will be explained in Section~\ref{sec:methods}.
We use an integer linear programming approach to
obtain an optimal hitting set and a well known greedy
algorithm~\cite{Vazirani-2001} to generate a near-optimal
solution.
Additional details regarding our approach appear
in Section~\ref{sec:methods}.

\medskip

\noindent
\textbf{Organization:} The remainder of this paper
is organized as follows.
Section~\ref{sec:prelim} provides formal definitions of
descriptors and the problem of constructing minimal
descriptors. It also includes examples to illustrate
the definitions.
Section~\ref{sec:methods} outlines the
methods used to generate the descriptors.
Section~\ref{sec:data_desc} describes the datasets used
in our work. 
Section~\ref{sec:experiments} examines the descriptors
obtained through our methods for three of the datasets
and provides a discussion on the 
insights obtained from the descriptors.
Section~\ref{sec:concl} presents conclusions,
limitations and
directions for future work.
Appendix~A discusses our results for an additional dataset
and Appendix~B includes some other outputs generated 
by our experiments.

                     
\section{Preliminaries} \label{sec:prelim}

\subsection{Basic Ideas}
\label{sse:basic_ideas}

We assume throughout this paper that a 
clustering algorithm \emph{partitions} a given dataset into  a specified number
$K \geq 2$ of clusters.
Several popular clustering algorithms (e.g., $k$-means
and its variants, DBSCAN, BIRCH) 
belong to the class of partitional clustering
algorithms (see e.g., \cite{Tan-etal-2019,Han-etal-2012,Zaki-etal-2020}).
We consider a \emph{post~hoc} explanation approach,
where the explanation phase begins after the clustering step has been completed.
Given a cluster consisting of a subset of data items, we want to explain why the clustering algorithm placed
those data items in the same cluster.
Following \cite{DGR-NIPS-2018},
we use \textbf{auxiliary} information (tags) for this purpose.
We now present a simple example to illustrate the use
of tags.

\noindent 
\textbf{Example:}~ Suppose we have a cluster of images. Possible tags for an
image are objects appearing in the image (e.g., baby, dog, bicycle, tree).
For a cluster of images, the following are 
examples of the two forms of explanations 
considered in this paper.

\begin{enumerate}
\item Each image contains at least one item
from the following set of objects:
\{baby, bicycle, dog\}.
This is an example of a  \textbf{Disjunctive} explanation.

\item Each image contains at least one item
from each of the following subsets of objects.

\smallskip
\hspace*{0.5in}
\{baby, dog\} ~\textbf{and}~ \{car, house\}.

\smallskip 

\noindent 
This is an example of a 
\textbf{Conjunctive Normal Form} (CNF) style \textbf{explanation}.
It contains two clauses, namely ``baby \textbf{or} dog'' and ``car \textbf{or} house''.
In other words,
this explanation specifies two sets, and each
image in a cluster contains at least one item from each of the two sets.
Thus, the CNF-style explanation provides more information about the
composition of a cluster.
\end{enumerate}
One can think of a disjunctive
explanation as a
CNF-style explanation that uses just
one clause.
Although one can consider CNF-style explanations
with three or more clauses, 
we restrict our attention
to CNF-style descriptions with two clauses
for two reasons.
First, a larger number of clauses may make it difficult 
for humans to understand the description.
Second, it would be computationally more
expensive to generate descriptions with more clauses.
In Section~\ref{sse:scale} we present
experimental results to support this observation.

\subsection{Formal Definitions}
\label{sse:formal_def}

We now formalize the ideas presented in 
Section~\ref{sse:basic_ideas} using the 
terminology from~\cite{DGR-NIPS-2018}.
Suppose a cluster $C$ = 
\{$d_1$, $d_2$, $\ldots$, $d_n$\}
consists of $n$ data items.
Let $U = \{t_1, t_2, \ldots, t_m\}$ denote the set of all possible tags.
We refer to $U$ as the \textbf{universe} of tags. 
Each data item $d_i$ is associated with 
a subset $T_i \subseteq U$ of tags that \textbf{describe} $d_i$, $1 \leq i \leq n$.


\begin{definition} \label{def:disj_desc}
A \textbf{valid disjunctive descriptor} $D$ for a cluster $C$ is
a subset of tags (i.e., $D \subseteq U$) such that
for each data item $d_i \in C$, $D$ contains at least one of the tags in $T_i$. Formally, 
for each $d_i \in D$, 
$D \cap T_i \neq \emptyset$.
\end{definition}

In general, a valid descriptor may contain \textit{too many} tags. 
Since shorter explanations are easier to understand,
it will be useful to find a descriptor $D$ with
a \emph{minimum} number of tags.

The formal definition of a CNF-style descriptor with 
two disjunctive sets is as follows.

\begin{definition}\label{def:cnf_desc}
A CNF-style descriptor
has the form ``$D_1$ \textbf{and} $D_2$'',
where $D_1$ and $D_2$ are subsets of tags that
satisfy
the following two conditions:
(i)~$D_1$ and $D_2$ are valid disjunctive 
descriptors for the cluster and
(ii)~sets $D_1$ and $D_2$ are \emph{disjoint};
  that is, $D_1 \cap D_2$ = $\emptyset$.
\end{definition}

The \textbf{size} of a CNF-style descriptor with two disjunctive
sets $D_1$ and $D_2$ is given by $|D_1| + |D_2|$.
As in the case of a disjunctive descriptor, it will be
desirable to minimize the size of a CNF-style descriptor.

We now present a simple example to illustrate
the above formal definitions.

\medskip

\noindent
\textbf{Example:}
Suppose cluster $C$ = \{$d_1$, $d_2$, $d_3$\}
has three data items.
Let $U = \{t_1, t_2, t_3, t_4, t_5, t_6\}$ denote the universe of tags and
let the respective tag sets $T_1$, $T_2$ and $T_3$
for the three data items $d_1$, $d_2$ and $d_3$ be as follows:
\begin{center}
$T_1 = \{t_1, t_2, t_5\},~~ T_2 = \{t_3, t_4, t_5\}~~ 
\mathrm{and}~~ T_3 = \{t_3, t_4, t_6\}.$
\end{center}
\begin{itemize}
\item Here, a valid descriptor $D_1$ for $C$ is 
$D_1$ = $\{t_1, t_4, t_6\}$.
This is because tags $t_1$, $t_4$ and $t_6$
describe data items $d_1$, $d_2$ and $d_3$
respectively.

\item However, $D_2$ = $\{t_1, t_5\}$ is \textbf{not} a valid descriptor for $C$ since none of the tags in
$D_2$ describes $d_3$.

\item A smaller valid descriptor $D_3$ for $C$ is: $D_3$ = $\{t_2, t_3\}$. Here, $t_2$ describes $d_1$ and
$t_3$ describes both $d_2$ and $d_3$.
As one can readily verify, no single tag describes
all the three data items in $C$.
Thus,
$D_3$ is a valid descriptor of \textit{minimum size}.

\item A CNF-style descriptor $D_4$ with two 
clauses for $C$ is:
$\{t_1, t_3\}$ \textbf{and} $\{t_5, t_6\}$.
As can be verified, each data item in $C$ is described
by  at least one tag in $\{t_1, t_3\}$ and 
at least one tag in $\{t_5, t_6\}$.
Note that the two subsets of tags 
included in the CNF-style descriptor $D_4$
are \emph{disjoint}.
It can also be verified that $D_4$ is of minimum size
among all CNF-style descriptors with two disjunctive subsets
of tags.
\end{itemize}

\noindent
\textbf{A note about CNF-style descriptors:}~ For some clusters,
a CNF-style descriptor with two disjunctive subsets may not
exist. To see this, suppose a cluster
$C = \{d_1, d_2, d_3, d_4\}$ has four data items.
Let $U = \{t_1, t_2, t_3, t_4, t_5\}$ denote the universe
of tags.
Let $T_1 = \{t_1\}$, $T_2 = \{t_2, t_3\}$,
$T_3 = \{t_3, t_4\}$ and $T_4 = \{t_4, t_5\}$ 
denote the descriptors of data items $d_1$, $d_2$, $d_3$
and $d_4$ respectively.
Since $t_1$ is the only tag that describes $d_1$, it must  appear in every valid descriptor for $C$.
However, since a CNF-style descriptor requires two \emph{disjoint} disjunctive descriptors for $C$, we conclude
that one cannot find a CNF-style descriptor for $C$.
We did not encounter this situation in our experiments. 
If this happens with some datasets, one may remove the disjointness condition and allow a small overlap between
the two disjunctive descriptors.

In the next section, we outline our methods
for generating disjunctive and CNF-style
descriptors.


\section{Methods Used to Generate Descriptors}
\label{sec:methods}

\subsection{Overview}\label{sse:overview}
In this section, we discuss our methods for generating
both disjunctive and CNF-style descriptors.
As will be seen, the disjunctive descriptor corresponds
to a well known set theoretic concept called
a \textbf{hitting set} \cite{GJ-1979}.
The CNF-style descriptor is an extension of
this concept.

\subsection{Constructing Disjunctive Descriptors}
\label{sse:disjunctive_alg}

\newcommand{\cala}{\mbox{$\mathcal{A}$}}

We begin with a formal definition of the concept
of a hitting set to point out how
it corresponds to the notion of a cluster descriptor.

\smallskip

\noindent
\begin{definition} \label{def:hitting_set}
(i) Suppose \cala{} = $\{A_1, A_2, \ldots, A_n\}$
is a collection of $n$ subsets of a universal set $U$.
A subset $X \subseteq U$ \textbf{hits} a subset $A_j \in \cala{}$ if $X \cap A_j \neq \emptyset$; that is,
$X$ has at least one element that appears in $A_j$.
A subset $H \subseteq U$ is a \textbf{hitting set} for \cala{} if it hits every subset $A_i \in \cala{}$;
formally, for each $A_i \in \cala{}$, 
$H \cap A_i \neq \emptyset$. 
(ii) A \textbf{minimum hitting set} for \cala{} is hitting set
with a \emph{minimum} number of elements.
\end{definition}

\medskip

\noindent
\textbf{Example:} Suppose $U = \{t_1, t_2, t_3, t_4, t_5, t_6\}$.
Consider the following collection \cala{} of four subsets of $U$:
\[
A_1 ~=~ \{t_1, t_3\},~~
A_2 ~=~ \{t_2, t_3, t_5\},~~
A_3 ~=~ \{t_4, t_5, t_6\}~~ \mathrm{and}~~
A_4 ~=~ \{t_2, t_4, t_6\}.
\]
The set $U$ = $\{t_1, t_2, t_3, t_4, t_5, t_6\}$ is a
\emph{trivial} hitting set for \cala{}.
The set $H_1 = \{t_1, t_5, t_6\}$ is a smaller hitting set for \cala{}.
Further, the set $H_2 = \{t_3, t_6\}$ is a \emph{minimum} hitting set for \cala{} since 
$H_2$ is a hitting set for \cala{} and no single 
element of $U$ is a hitting set for \cala{}.

\medskip

\noindent
\textbf{Relationship between hitting sets and cluster
descriptors:}~
Let us compare the definitions of a valid descriptor
(Definition~\ref{def:disj_desc}) and that of a
hitting set (Definition~\ref{def:hitting_set}).
In the hitting set definition, by thinking of the universe of tags as the set $U$ 
and the collection
of tag sets for the data items of a cluster as the subset collection \cala{},
it can be seen that
each valid disjunctive descriptor for the cluster is
a hitting set for \cala.

\medskip

In general, the problem of computing a minimum hitting set is \cnp-hard~\cite{GJ-1979}.
So, We don't expect the problem to be efficiently solvable
in general. When $|U|$ and $|\cala|$ are large, 
we will use an efficient heuristic to compute near-minimal hitting sets.
For reasonably small problem instances, one can compute minimum hitting sets
in a reasonable amount of time using an
integer linear programming (ILP) formulation.
We will discuss the heuristic and the ILP approaches next.


\subsection{A Heuristic for Minimum Hitting Set (MHS)}

A well known greedy heuristic for finding an MHS 
(see e.g., \cite{Vazirani-2001}) for a collection of
subsets of a base set $U$
is outlined in Figure~\ref{fig:mhs_heur}.
The basic idea is to iteratively choose a tag from
$U$ that hits the largest number of 
remaining tag sets.
The heuristic ends when all the tag sets have been hit.
By the correspondence between hitting sets
and disjunctive descriptors, the resulting hitting
set is also a valid disjunctive descriptor for the
cluster.

\begin{figure}[t]
\noindent
\rule{\textwidth}{0.01in}

\smallskip

\noindent
\textbf{Input:}~ A universal set $U$, a collection \cala{} of
subsets of $U$.
($U$ corresponds to the universe of tags and \cala{}
represents the collection of tag sets corresponding to
the elements of a cluster.)

\smallskip

\noindent
\textbf{Output:}~ A (small) hitting set $H$ for \cala{}.
(This hitting set serves as a disjunctive descriptor for
the cluster.)

\smallskip

\noindent
\textbf{Steps of the Algorithm:}

\begin{enumerate}
\item Let $S$ = \cala{} (the sets which have \emph{not}
yet been hit) and $H ~= \emptyset$.
($H$ will be the hitting set produced by the algorithm.)

\item For each tag $t_i \in U$, compute the frequency $f_i$,
that is, the
number of sets in $S$ in which $t_i$ occurs. 

\item \textbf{while}~ ($S \neq \emptyset$)
  \begin{description}
    \item{(i)} Choose a tag $t_j$ with the maximum frequency value $f_j$. 
    \item{(ii)} Add $t_j$ to $H$ and delete $t_j$ from $U$. 
    \item{(iii)} Delete from $S$ all sets that contain tag $t_j$. 
    \item{(iv)} For each tag $t_i \in U$, revise the frequency
     count $f_i$. 
  \end{description}

\item Output $H$. (This is the descriptor returned for
the cluster whose elements have the tag sets given by \cala{}.)
\end{enumerate}
\caption{A Greedy Heuristic for Finding a Small Hitting Set (i.e., Disjunctive Descriptor)}
\label{fig:mhs_heur}
\smallskip
\noindent
\rule{\textwidth}{0.01in}
\end{figure}


\medskip

\noindent
{\textbf{An Example for the Minimum Hitting Set Heuristic:}}

\smallskip

\noindent
\underline{\textsf{(i) Initially:}} $U = \{t_1, t_2, t_3, t_4, t_5, t_6\}$.
\cala{} has the following 4  subsets of $U$:
$T_1 ~=~ \{t_1, t_3\}$,\\  
$T_2 ~=~ \{t_2, t_3, t_5\}$,
$T_3 ~=~ \{t_4, t_5, t_6\}$,~ and~ 
$T_4 ~=~ \{t_2, t_4, t_6\}$.

\smallskip

Also, $S = \{A_1, A_2, A_3, A_4\}$, $H = \emptyset$,
$f_1 = 1$, $f_2 = f_3 = f_4 = f_5 = f_6 = 2$.
(As mentioned in Figure~\ref{fig:mhs_heur}, $f_i$ represents the number of tag sets
which tag $t_i$ appears, $i = 1, 2, 3, 4$.)

\smallskip

\noindent
\underline{\textsf{(ii) Iteration 1:}} Choose $t_2$ (which has the
maximum frequency of 2). Then $H = \{t_2\}$,
$U = \{t_1, t_3, t_4, t_5, t_6\}$, $S = \{A_1, A_3\}$,
$f_1 = f_3 = f_4 = f_5 = f_6 = 1$.

\smallskip

\noindent
\underline{\textsf{(iii) Iteration 2:}} Choose $t_1$. Then $H = \{t_1, t_2\}$,
$U = \{t_3, t_4, t_5, t_6\}$, $S = \{A_3\}$,
$f_3 = 0$, $f_4 = f_5 = f_6 = 1$.

\smallskip

\noindent
\underline{\textsf{(iv) Iteration 3:}} Choose $t_4$. Then $H = \{t_1, t_2, t_4\}$,
$U = \{t_3, t_5, t_6\}$, $S = \emptyset$.

\smallskip

\noindent
\underline{\textsf{Output of the algorithm:}}~ 
$H = \{t_1, t_2, t_4\}$.

\medskip

\noindent
\textbf{Note:}~ The set $H$ is indeed a hitting set for the
set collection \cala{}. However, it is \emph{not} a minimum hitting
set. One minimum hitting set 
for this example is $H_1 = \{t_3, t_6\}$.

\subsection{An Integer Linear Program for Obtaining
a Minimum Hitting Set}\label{sse:ilp_for_mhs}

We now present an integer linear 
program (ILP) for the minimum hitting set problem.
As mentioned earlier, the resulting minimum hitting set
serves as a disjunctive descriptor.
As before let $U = \{t_1, t_2, \ldots, t_m\}$ denote the
universe of tags and let 
\cala{} = $\{T_1, T_2, \ldots, T_n\}$ denote the tag
subsets of the $n$ data items in a cluster $C$.

\smallskip

\noindent
\underline{\textsf{Variables:}} For each tag $t_i \in U$,
we have a $\{0,1\}$-valued variable $x_i$, $1 \leq i \leq m$.
The significance of these variables is as follows:
tag $t_i$ appears in a minimum hitting set if and only
if $x_i$ has the value 1.

\smallskip

\noindent
\underline{\textsf{Objective:}}~ \textsc{Minimize}~
$\sum_{i=1}^{m} x_i$.

\smallskip

\noindent
\underline{\textsf{Constraints:}}

\begin{enumerate}
\item For each tag set $T_j \in \cala{}$, we need to
ensure that the chosen hitting set contains at least
one of the tags in $T_j$.
The corresponding set of constraints is as follows:
\begin{equation}\label{eqn:occurrence}
\sum_{t_i\,\in\, T_j} x_i ~\geq~ 1, ~~1 \leq j \leq n.
\end{equation}

\item $x_i \in \{0,1\}$,~ $1 \leq i \leq m$.
\end{enumerate}

\noindent
\underline{\textsf{Obtaining a minimum hitting set:}}~
For $1 \leq i \leq m$, choose each tag $t_i$ for
which $x_i = 1$.

\smallskip 

In our experiments, we used the 
Gurobi solver~\cite{gurobi-2023}
to obtain solutions to ILPs.


\subsection{A Simple Approach for Generating a CNF Descriptor With Two Clauses}

Recall that a CNF descriptor with two clauses has the form
``$D_1$ \textbf{and} $D_2$'', where $D_1$ and $D_2$ are both valid
disjunctive descriptors for a cluster.
We use the following simple approach for generating a DNF descriptor with two clauses.

\begin{enumerate}
\item Suppose there are $n$ data items.
For a data item $d_i$, let $T_i$ denote its tag set, $1 \leq i \leq n$.

\item For the tag sets $T_1$, $T_2$, $\ldots$, $T_n$,
find a descriptor $D_1$ using the ILP or the heuristic
for finding a small  hitting set.

\item From each tag set $T_i$, remove the tags in $D_1$.
(Formally, replace each tag set $T_i$ by the set $T_i - D_1$, $1 \leq i \leq m$.)

\item For the modified tag sets from Step~3, find a
descriptor $D_2$ using the ILP or the heuristic
for finding a small  hitting set.

\item Output ``$D_1$ \textbf{and} $D_2$'' as the
CNF descriptor (with two clauses).
\end{enumerate}

\medskip

\noindent
\textbf{Example:} We will now illustrate the above approach
for finding a CNF descriptor with two clauses.

\begin{enumerate}
\item Suppose a cluster $C$ has three data items $d_1$, $d_2$, $d_3$.
Let $U = \{t_1, t_2, t_3, t_4, t_5, t_6\}$ denote the universe of tags.
Let the tag sets for these items be as follows:
\begin{center}
$T_1 = \{t_1, t_2, t_5\},~~ T_2 = \{t_3, t_4, t_5\}~~ 
\mathrm{and}~~ T_3 = \{t_3, t_4, t_6\}.$
\end{center}

\item A hitting set for $T_1$, $T_2$ and $T_3$ is $D_1 = \{t_1, t_3\}$.
(The ILP-based approach can find this hitting set; the heuristic may
find a slightly larger hitting set.) \smallskip

\item When we remove the tags in $D_1$ from each of the three tag sets,
the new tag sets are:
\begin{center}
$T_1 = \{t_2, t_5\},~~ T_2 = \{t_4, t_5\}~~ 
\mathrm{and}~~ T_3 = \{t_4, t_6\}.$
\end{center}

\item A hitting set for the (new) tag sets $T_1$, $T_2$ and $T_3$
is $D_2 = \{t_5, t_6\}$. (This can also be found using the ILP-based
approach or the heuristic.) \smallskip

\item Thus, a CNF descriptor is
``$\{t_1, t_3\}$ \textbf{and} $\{t_5, t_6\}$''.
\end{enumerate}

\subsection{Limitations of the Approach for Generating CNF-Style Descriptors}
\label{sse:limitations}

\begin{enumerate}
\item If the tag set for some data item in the given cluster $C$
contains only one tag,
a CNF descriptor with two (or more) clauses does not exist
for the cluster $C$.
So, before attempting to construct a CNF descriptor, all
data items whose tag sets have only one tag must be removed from $C$
to obtain a new cluster $C'$ (for which one can try to find
a CNF descriptor).

\item After finding the first descriptor $D_1$, note that Step~3
of the approach for finding a CNF descriptor eliminates the tags in
$D_1$ from each tag set.
If this step causes the tag set $T_i$ of data item $d_i$ to become empty, once
again the approach cannot find a CNF style descriptor for the cluster $C$.
Thus, all such items whose tag sets become empty in Step~3 should also be
removed from $C$ to obtain a new cluster $C'$
(for which one can try to find a CNF descriptor).
\end{enumerate}


\begin{figure}[tbh]
\centering
\begin{minipage}{.47\textwidth}
  \hspace*{-0.1in}
  \includegraphics[width=7cm]{{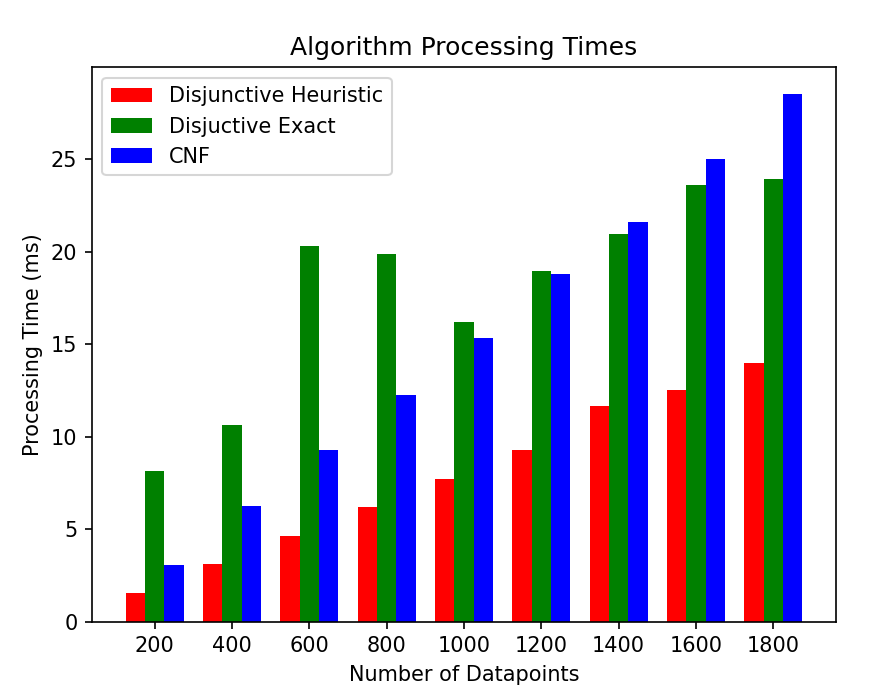}}
  \captionof{figure}{Algorithm Execution Times for 2000 Datapoints}
  \label{fig:ex_times_2000}
\end{minipage}%
\quad 
\begin{minipage}{.47\textwidth}
  \includegraphics[width=7cm]{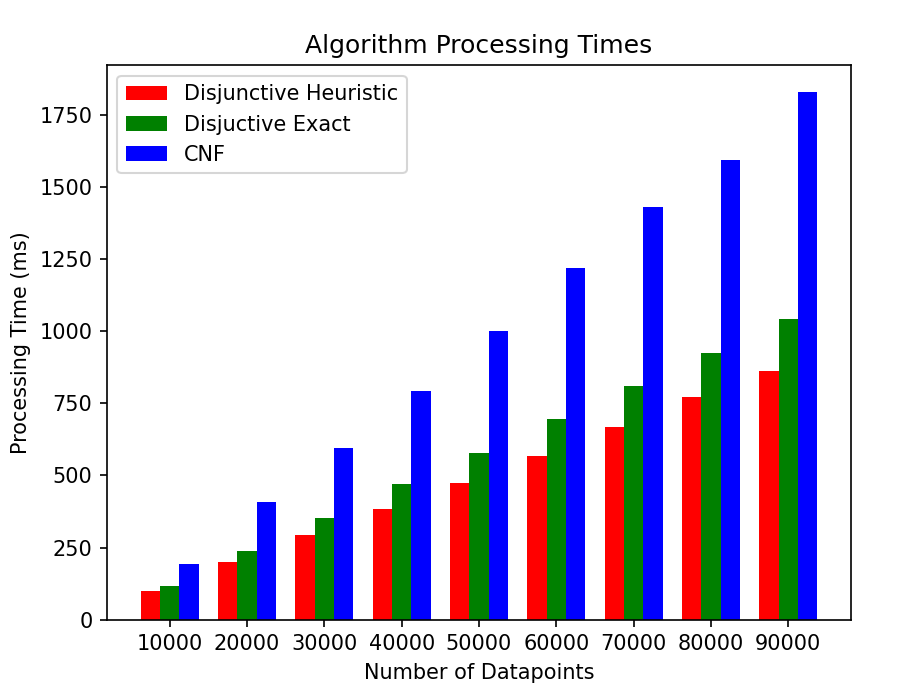}
  \captionof{figure}{Algorithm Execution Times for 100,000 Datapoints}
  \label{fig:ex_times_100000}
\end{minipage}
\end{figure}

\subsection{Scalability of the Methods}
\label{sse:scale}

To get an idea regarding the scalability of the
explanation methods discussed in this section, we carried out an experimental study using synthetic datasets.
Specifically, we created synthetic datasets whose
sizes varied from 200 to 100,000 and measured the running times of the exact and heuristic algorithms for
generating disjunctive descriptors and the one for
generating CNF descriptors.
The experiments were carried out on a machine that uses a  
high-performance Intel Xeon Gold 6230 CPU @ 2.10GHz with 384 GB of memory and a Rocky Linux~8 operating system. These timing studies were run on a single core using Python 3.10.9 scripts. The experiments involved constructing synthetic datasets and taking the average of 10 time calculations for the designated number of 
data points of each iteration.
The results are shown in Figures~\ref{fig:ex_times_2000}
and ~\ref{fig:ex_times_100000}.
As can be seen from these figures,
the disjunctive exact algorithm performs slower on smaller datasets, highlighting an even greater advantage of heuristic solvers for these cases. For larger datasets, the CNF algorithm takes roughly twice as long as the disjunctive solvers, as the disjunctive approach is applied twice to generate the two clauses required 
for the CNF.  
The increase in time for generating CNF explanations
becomes more pronounced as the size of the dataset
is increased.
This justifies our choice of limiting the CNF descriptor
to two clauses.

\section{Descriptions of Datasets Used}
\label{sec:data_desc}

\subsection{Overview}
In this section, we provide descriptions of the four datasets used
in our work, three of which will be covered in the main body of the
paper, with an additional dataset examined in Appendix~A.  Some
information about these datasets appears in
Table~\ref{tab:experimental_datasets}.  The datasets vary in terms
of size, number of clusters, and number of tags defined for each
data point and they span different domains, including education,
entertainment, psychology, and demographics.
This provides a comprehensive evaluation of our explanation method
across multiple contexts. The clustering and tagging process for
each dataset is discussed in detail in Section~\ref{sec:experiments}.

\begin{table}[htbp]
  \centering
  \begin{tabular}{|c|c|c|c|p{1.6in}|}
    \hline
    \textbf{Dataset} & \textbf{Size} 
         & \textbf{\#Clusters} & \textbf{\#Tags} 
         & \textbf{Description} \\
    \hline\hline
    College Majors & 173 & 3 & 30 & {Graduate Major Statistics from 2010-12 Survey}\\
    \hline
    Movies & 638 & 4 & 19 & {Most Popular Movie\newline Releases 1989--2021}\\
    \hline
    Divorce Predictors & 170 & 2 & 108 & {Married Couple's \newline Survey Responses} \\
    \hline
    1994 Census Income & 30,717 & 2 & 29 & {1994 Adult Census\newline Information}\\
    \hline
  \end{tabular}
  \caption{Real-World Datasets}
  \label{tab:experimental_datasets}
\end{table}

\subsection{College Majors Dataset}
The College Majors dataset \cite{misc_college_majors} compiles information on various college graduate school majors from the 2010-2012 American Community Survey. Comprising 173 distinct majors, the dataset offers 22 features concerning graduates of each major (refer to Table \ref{tab:college_data_original_features}). Details regarding the tag definition and application process for this dataset are expounded upon in the Experiments section.

\begin{table}[ht]
\centering
\scriptsize
\begin{tabular}{@{}ll@{}}
\toprule
\textbf{Feature Name}             & \textbf{Description}                       \\ \midrule
\texttt{Major\_code}              & Code identifying the major                 \\
\texttt{Major}                    & Name of the major                          \\
\texttt{Major\_category}          & Category of the major                      \\
\texttt{Grad\_total}              & Total number of graduates                  \\
\texttt{Grad\_sample\_size}       & Sample size of graduates                   \\
\texttt{Grad\_employed}           & Number of employed graduates               \\
\texttt{Grad\_full\_time\_year\_round} & Graduates employed full-time, year-round \\
\texttt{Grad\_unemployed}         & Number of unemployed graduates             \\
\texttt{Grad\_unemployment\_rate} & Unemployment rate among graduates          \\
\texttt{Grad\_median}             & Median salary for graduates                \\
\texttt{Grad\_P25}                & 25th percentile salary for graduates       \\
\texttt{Grad\_P75}                & 75th percentile salary for graduates       \\
\texttt{Nongrad\_total}           & Total number of non-graduates              \\
\texttt{Nongrad\_employed}        & Number of employed non-graduates           \\
\texttt{Nongrad\_full\_time\_year\_round} & Non-graduates employed full-time, year-round \\
\texttt{Nongrad\_unemployed}      & Number of unemployed non-graduates         \\
\texttt{Nongrad\_unemployment\_rate} & Unemployment rate among non-graduates    \\
\texttt{Nongrad\_median}          & Median salary for non-graduates            \\
\texttt{Nongrad\_P25}             & 25th percentile salary for non-graduates   \\
\texttt{Nongrad\_P75}             & 75th percentile salary for non-graduates   \\
\texttt{Grad\_share}              & Share of graduates within the group        \\
\texttt{Grad\_premium}            & Salary premium for graduates over non-graduates \\
\bottomrule
\end{tabular}
\caption{College Majors Dataset Original Features}
\label{tab:college_data_original_features}
\end{table}

\subsection{Movies Dataset} 
The Movies dataset \cite{misc_movies_dataset} encompasses 636 of the most popular films released between 1989 and 2021, characterized by eleven distinct features. For the purpose of clustering, the k-means algorithm was applied using the following attributes: title, MPAA rating, budget, gross, release date, genre, runtime, rating, and rating count. To facilitate the interpretation of the clustering outcomes, tags were assigned to these features. The methodology for tagging these features is detailed in the Experiments section.

\subsection{Divorce Predictors Dataset} 
The Divorce Predictors dataset was obtained from the UC Irvine Machine Learning Repository \cite{miscDivorcePredictorsDataSet}. This dataset was originally derived from a survey consisting of 54 questions administered to 170 couples, based on the Gottman Methods approach to couples therapy. This approach aims to mitigate conflicting verbal communication by providing a comprehensive assessment of a couple's relationship \cite{gottman_method_website}. The dataset includes a data file where each row represents one of the 170 couples, and each column corresponds to one of the 54 survey questions. Participants rated their agreement with each statement on a scale from 0 to 4, with 0 indicating complete disagreement and 4 indicating complete agreement.

\subsection{Census Dataset}
The Census Income dataset \cite{misc_adult_2}, comprises data from 30,717 adults drawn from the 1994 Census Bureau database, including various demographic and social attributes. The dataset features the following attributes: age, work class, education, years of education, hours worked per week, and native country. The process of defining tags for these features is detailed in the Experiments section.

\section{Experiments}
\label{sec:experiments}

\subsection{Overview}

The general process for each experiment involves first preprocessing
the dataset to ensure its suitability for clustering. This includes
cleaning the data, creating dummy variables for categorical data,
and excluding irrelevant or problematic features (e.g., features
with missing or noisy values).  The next step is determining the
optimal number of clusters using the elbow method. Once the number
of clusters is established, a clustering algorithm is applied to
the data. Tags are then defined for each dataset based on the
features of the dataset. After applying the tags to each datapoint
in the dataset, the hitting set solvers are applied and the tags
in the hitting sets are used to identify common characteristics or
patterns within each cluster. These tags are used to interpret the
reasons behind the clustering, providing insights into the common
attributes of the data points in each group. Visualization techniques,
such as principal component analysis, are used to display the
clustering results and facilitate further analysis. In the following
subsections, the above process is applied to three different datasets.
(Our results for the Census dataset can be found in Appendix~A.)
The resulting hitting sets are then examined to determine the
practical implications of the explanations. 

\subsection{College Majors}\label{sse:expt_majors}

This experiment aims to cluster graduate school majors using features
sourced from the dataset provided by FiveThirtyEight on
\texttt{data.world}. By identifying commonalities among majors in
a cluster using tags, inferences can be made about the reasons
behind the clustering. To achieve this, the dataset underwent
preprocessing, including the creation of dummy variables for
categorical data. Additionally, features that were likely to
contribute to multicollinearity or those that were deemed irrelevant
for clustering purposes, such as `Major code', `Grad unemployment
rate', `Nongrad unemployment rate', `Grad median', `Nongrad median',
and `Grad share', were excluded. The preprocessed dataset was used
in subsequent procedures.

Utilizing the $k$-means clustering algorithm, clusters were generated
based on the aforementioned features. Employing the elbow method,
the optimal partitioning for the data was determined to be 3 clusters.

\begin{figure}
\centering
\begin{minipage}{.47\textwidth}
  \hspace*{-0.1in}
  \includegraphics[width=7cm]{{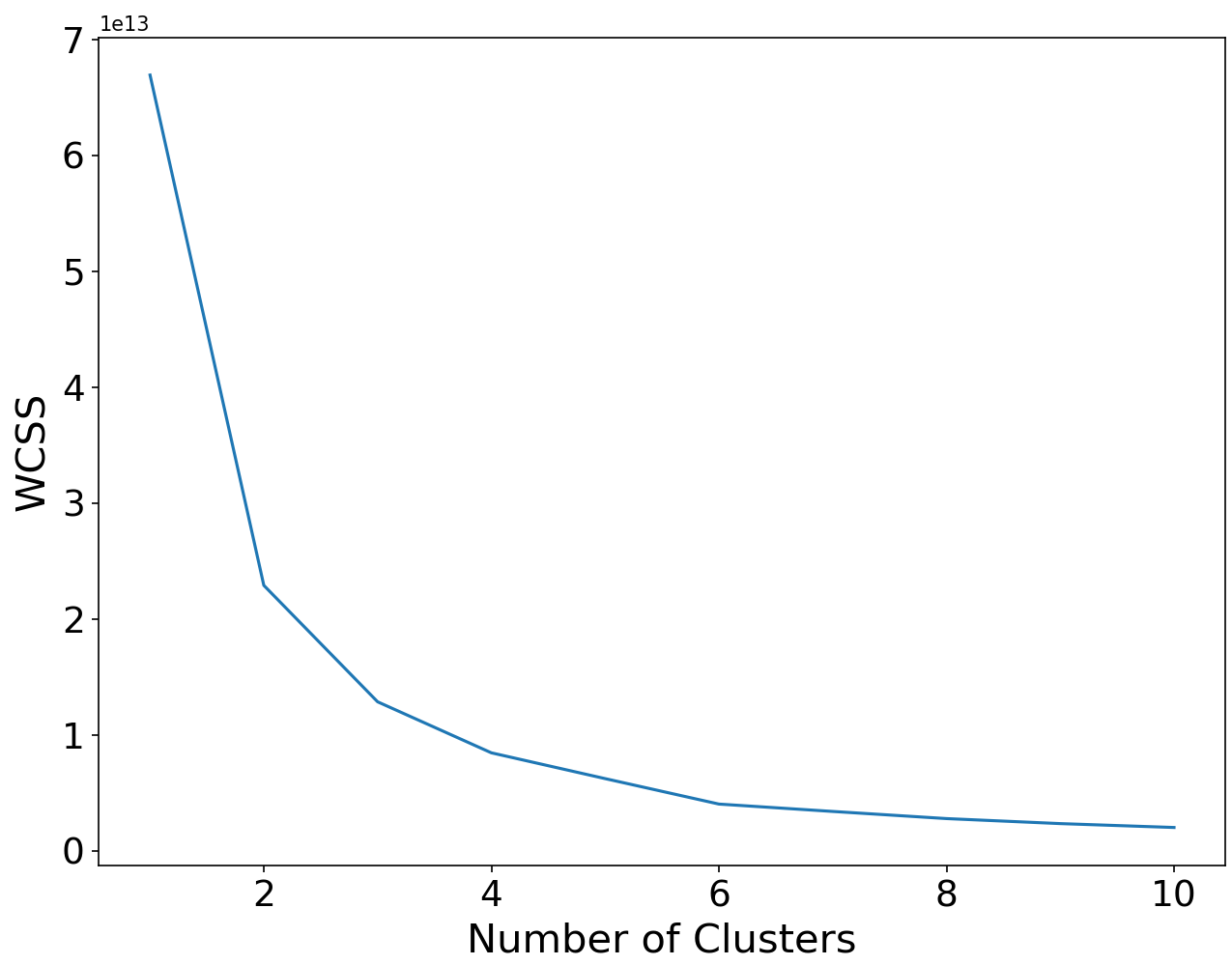}}
  \captionof{figure}{College Majors Data: Elbow Method.}
  \label{fig:college_elbow}
\end{minipage}%
\quad 
\begin{minipage}{.47\textwidth}
  \includegraphics[width=7cm]{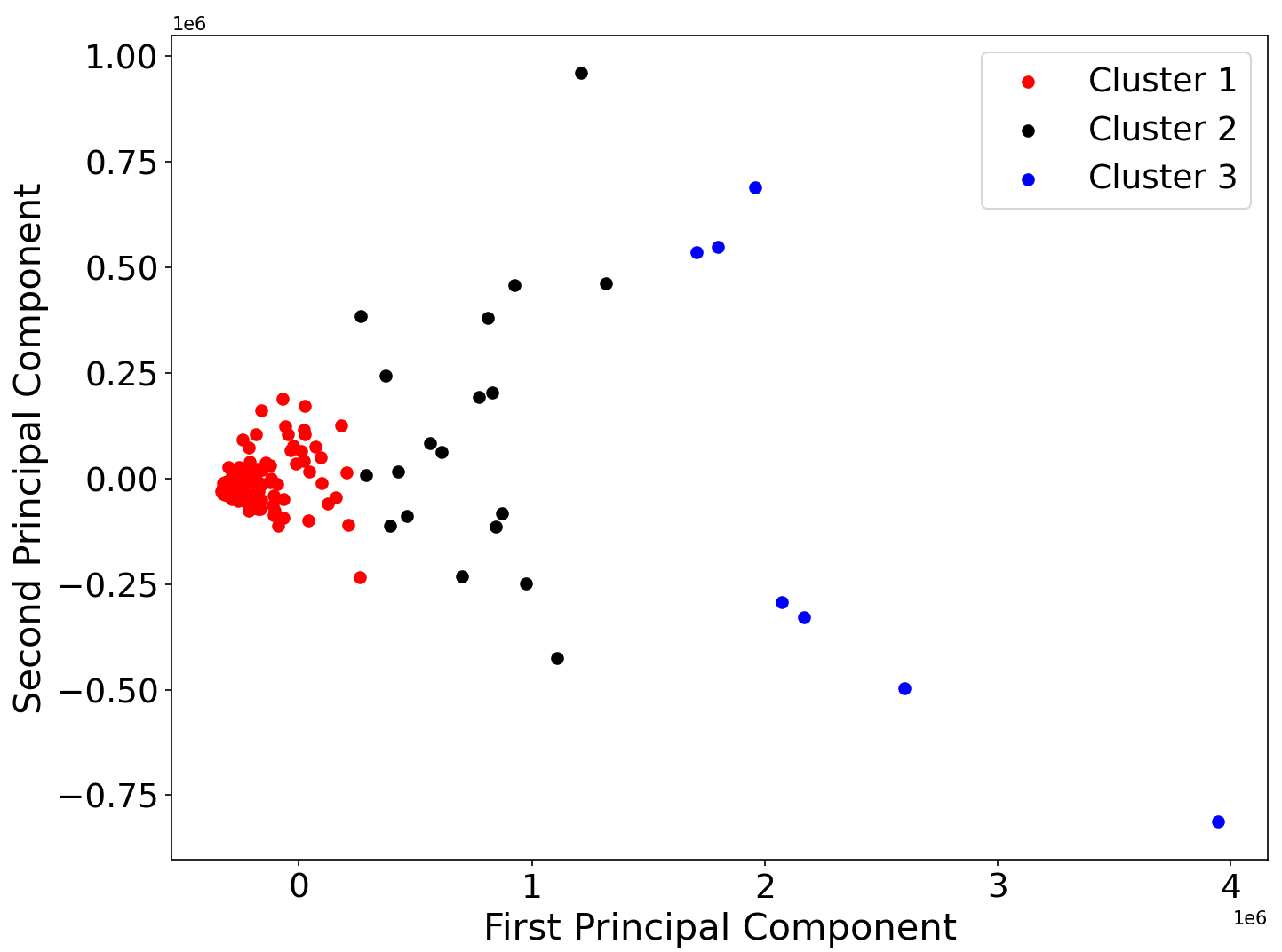}
  \captionof{figure}{College Majors Data: Principal Component Analysis.}
  \label{fig:college_pca}
\end{minipage}
\end{figure}

The $k$-means algorithm was applied to the data to generate three
clusters.  In the resulting clustering, the majors were distributed
among the three clusters in an intriguing manner, with one cluster
notably larger than the other two. From the PCA results
(Figure~\ref{fig:college_pca}), it would be reasonable to conclude
that the majors in Cluster 3 are outliers and represent the ``most
different'' majors based on the provided features.

The tags were defined for each datapoint as being below or above
the median value for each feature. Therefore, each feature used for
clustering has two associated tags, which are complementary. The
tag definitions are provided in Table~\ref{fig:college_Tags}.

\begin{longtable}{|>{\centering\arraybackslash}m{0.1\textwidth}|>{\centering\arraybackslash}m{0.7\textwidth}|}
\hline
\textbf{Tags} & \textbf{Attribute} \\ \hline\hline
\endfirsthead
\hline
\textbf{Tags} & \textbf{Attribute} \\ \hline\hline
\endhead
t1 / t2 & Major category in arts and sciences OR mathematics and engineering \\ \hline
t3 / t4 & Total graduates in major below OR above median \\ \hline
t5 / t6 & Graduate sample size for major below OR above median \\ \hline
t7 / t8 & Graduates employed for major below OR above median \\ \hline
t9 / t10 & Graduates from major working full-time year-round below OR above median \\ \hline
t11 / t12 & Graduates from major unemployed below OR above median \\ \hline
t13 / t14 & Number of graduates in major in top 25\% of class below OR above median \\ \hline
t15 / t16 & Number of graduates in major in top 75\% of class below OR above median \\ \hline
t17 / t18 & Number of people in major who did not graduate below OR above median \\ \hline
t19 / t20 & Number of people in major who did not graduate and are employed below OR above median \\ \hline
t21 / t22 & Number of people in major who did not graduate and work full-time year-round below OR above median \\ \hline
t23 / t24 & Number of people in major who did not graduate and are unemployed below OR above median \\ \hline
t25 / t26 & Number of people from major who did not graduate but were in top 25\% of class below OR above median \\ \hline
t27 / t28 & Number of people from major who did not graduate but were in top 75\% of class below OR above median \\ \hline
t29 / t30 & Graduate earnings premium for major below OR above median \\ \hline
\caption{College Majors Data Tag Definitions}
\label{fig:college_Tags}
\end{longtable}

Tags t1 and t2 served as a generalization of the 173 college majors provided in the original dataset. By abstracting the features, there is an increased likelihood of obtaining meaningful explanations for each cluster. This is because the hitting sets utilize tags to identify patterns among the features of the datapoints within each cluster. The broader the tag definitions, the greater the probability that a significant portion of the datapoints in the cluster will share common tags, facilitating the emergence of patterns through the hitting sets. For tags t1-t2, each of the 173 college majors falls into two distinct categories (see Table \ref{tab:t1_and_t2_supplementary}).

\begin{table}[ht]
\centering
\scriptsize
\begin{tabular}{|p{5cm}|p{9cm}|}
\hline
\textbf{Category}                               & \textbf{Majors} \\ \hline\hline
\texttt{Arts and Sciences Majors (t1)}          & \begin{tabular}[t]{@{}l@{}} 
    Industrial Arts \& Consumer Services, Arts, Business, \\ 
    Law \& Public Policy, Agriculture \& Natural Resources, \\ 
    Communications \& Journalism, Social Science, Health, \\ 
    Interdisciplinary, Physical Sciences, Humanities \& Liberal Arts, \\ 
    Psychology \& Social Work, Biology \& Life Science, Education 
\end{tabular} \\ \hline
\texttt{Engineering, Mathematics and Computer Science Majors (t2)} 
    & Engineering, Computers \& Mathematics \\ \hline
\end{tabular}
\caption{College Majors Data Tags t1 and t2 Supplementary Information}
\label{tab:t1_and_t2_supplementary}
\end{table}

The 30 tags mentioned above were subsequently applied to the data, yielding an ILP matrix suitable for input into the three hitting set solver algorithms (disjunctive heuristic, disjunctive exact, and CNF heuristic). These algorithms were employed to obtain descriptors for each of the three clusters. The results are given in Table \ref{table:college_majors_hitting_sets}.

\begin{table}[h!]
\centering
\begin{tabular}{|>{\centering\arraybackslash}m{3cm}|>{\centering\arraybackslash}m{3cm}|>{\centering\arraybackslash}m{3cm}|>{\centering\arraybackslash}m{2.7cm}|}
\hline
\textbf{} & {\textbf{Cluster 1:}\newline \textbf{148 majors}} & {\textbf{Cluster 2:}\newline \textbf{20 majors}} & {\textbf{Cluster 3:}\newline \textbf{8 majors}}\\ \hline\hline
\textbf{Disjunctive Heuristic} & [t1, t3, t17, t14, t15] & [t4] & [t1] \\ \hline
\textbf{Disjunctive Exact} & [t29, t30] & [t6] & [t6] \\ \hline
\textbf{CNF} & ([t1, t3, t17, t14, t15], [t5, t7, t11, t19, t21, t28, t26, t29, t25]) & ([t4], [t6]) & ([t1], [t4]) \\ \hline
\end{tabular}
\caption{College Majors: Hitting Sets}
\label{table:college_majors_hitting_sets}
\end{table}

Using the median for feature-based tag definitions can lead to issues, as some disjunctive exact hitting sets include complementary tags (e.g., t29 and t30, indicating whether a major's graduate earnings premium is below or above the median value for all majors). Hitting set solvers, similar to the disjunctive exact solver, attempt to find a hitting set of minimum size such that all data in a cluster can be explained by the fewest number of tags. However, all datapoints in a cluster can be explained by tags indicating that the cluster’s contents either fall below or above some value for a feature, resulting in trivial hitting sets like the disjunctive exact hitting set for Cluster 1.

To address this issue, a hitting set solver algorithm was developed to filter which tags can be considered for hitting set generation. This algorithm is straightforward: it takes an ILP formulation of a cluster and calculates the percentage of data explained by each tag. For two complementary tags\footnote{Note that all tags in the college majors dataset are complementary pairs, such as t1 and t2, t3 and t4, t5 and t6, etc.}, only the tag accounting for a greater percentage of the data is considered for hitting set generation. This is achieved by setting the percentage value for the complementary tag with the lower percentage to 0 before applying the normal hitting set solver functions. The hitting sets produced by this filter applied to the disjunctive exact solver are displayed in Table \ref{tab:NonCompResults}

\begin{table}[h!]
    \centering
    \begin{tabular}{| m{3cm} | m{7cm} |}
        \hline
        \textbf{Clusters} & \textbf{Disjunctive Exact Filtered for Non-Complementarity} \\ \hline\hline
        Cluster 1 & \texttt{[t3, t14, t15]} \\ \hline
        Cluster 2 & \texttt{[t4]} \\ \hline
        Cluster 3 & \texttt{[t8]} \\ \hline
    \end{tabular}
    \caption{Disjunctive Exact Filtered for Non-Complementarity}
    \label{tab:NonCompResults}
\end{table}

\begin{table}[h!]
\centering
\small 
\begin{tabular}{|>{\centering\arraybackslash}m{2cm}|>{\centering\arraybackslash}m{6cm}|>{\centering\arraybackslash}m{3cm}|>{\centering\arraybackslash}m{3cm}|}
\hline
\textbf{Cluster} & \textbf{Criteria for Majors in Cluster} & \textbf{Interpretation} & \textbf{Example Major} \\
\hline\hline
Cluster 1 & 
\begin{itemize}
    \item Total number of graduates is below the median for these majors
    \item Number of graduates who are in the top 25\% of their class is above the median for these majors
    \item Number of graduates who are in the top 75\% of their class is below the median for these majors
\end{itemize} & 
These are majors with few students who make up the higher percentiles of GPAs across all universities in the study. & 
English Literature \\
\hline
Cluster 2 & 
\begin{itemize}
    \item Total number of graduates is above the median for these majors
\end{itemize} & 
Majors with many graduates. & 
Business Administration \\
\hline
Cluster 3 & 
\begin{itemize}
    \item Number of graduates employed is above the median for these majors
\end{itemize} & 
Majors with high job placement. & 
Engineering \\
\hline
\end{tabular}
\caption{Classification of Majors into Clusters Based on Hitting Set Explanations}
\end{table}

The clustering of these majors and the associated explanations can provide practical guidance for high school and college academic advisors. Advisors can leverage these classifications to better support students in making informed decisions about their majors that align with their academic capabilities and career ambitions. For instance, majors in Cluster~1, which represent the vast majority of majors in the dataset, exhibit smaller student populations, as indicated by tag t3. Additionally, a significant portion of students in Cluster 1 are in the top 25\% of their class. Over half of the majors in Cluster 1 have more students than not with GPAs falling within the 25th percentile for all majors at their school, as evidenced by the 53.061\% prevalence of tag t14 in Cluster 1 (see Table \ref{tab:majors_tag_percentages} in Appendix B). Examples of majors in Cluster 1 include Forestry, Oceanography, Statistics and Decision Science, Counseling Psychology, Journalism, and Engineering Technologies. It can be inferred that the average students in Cluster 1 generally have higher GPAs because the majors in this cluster, on average, are less demanding than those in the other two clusters.

Cluster 2 can be entirely explained by tag t4, indicating that all majors in Cluster 2 have a total number of graduates above the median. This suggests the popularity and broad appeal of majors in this cluster. Examples of majors in Cluster 2 include Marketing and Marketing Research, Criminal Justice and Fire Protection, Communications, History, Mathematics, Political Science and Government, Biology, Chemistry, Finance, Computer Science, and various types of engineering. The content of the cluster aligns with expectations from the hitting set explanation.

Like Cluster 2, Cluster 3 can be entirely explained by one tag: t1. However, t1 is already part of Cluster 1's hitting set, providing no discriminatory information for Cluster 3. To address this, a new hitting set was generated using the non-complementarity filter described above (see Table \ref{tab:NonCompResults}). This new hitting set indicates that all majors in Cluster 3 contain tag t8, indicating graduate employment above the median for all majors. Examples of majors in Cluster 3 include Business, Nursing, Accounting, various education majors, and Psychology. These majors typically have high job placement rates and suggest strong connections with industry and high demand for graduates.

In essence, these hitting sets provide classifications for clusters formed by grouping majors together based on their similar graduate statistics. These classifications can then be used by academic advisors, students, and educators to support students in making informed decisions that align with their abilities and ambitions while also informing them of the realities of the job market. This clustering and classification method using hitting sets could ensure a more statistical and strategic academic advising process.

It is worth noting that the necessity of adjusting Cluster 3's hitting set with a filter highlights an interesting idiosyncrasy of the hitting set solvers. Multiple tags in Cluster 3 account for 100\% of the majors in the cluster. For instance, t4, t6, t8, t10, t18, t20, t22, and t24 are all tied to all majors in Cluster 3 (see Table~\ref{tab:majors_tag_percentages}
in Appendix~B). The disjunctive heuristic selects tags for the hitting set based on their highest frequency sequentially, selecting t4 for the hitting set since it was the first tag with the highest frequency of 100\%. On the other hand, the disjunctive exact solver, which uses Gurobi to find a hitting set of minimum size, employs a different process to derive the hitting set, resulting in a different tag selection.

\subsection{Movies}\label{sse:movies}
In this experiment, movies are clustered based on features obtained from James Gaskin’s Movies dataset on \texttt{data.world}. Tags are used to identify common characteristics among movies within the same cluster, providing insights into why these movies were grouped together (see Table~\ref{tab:movie_tags}).

\begin{table}[htbp]
 \bigskip
  \centering

  \begin{tabular}{ll}
    \toprule
    \textbf{Criteria} & \textbf{Description} \\
    \midrule
        MPAA rating: \\
            & t1 - General audiences (G or PG) \\
            & t2 - Mature audiences (PG-13 or R) \\
        Budget (USD): \\
            & t3 - Below median: 60,000,000 - 80,000,000 \\
            & t4 - Above median: 80,000,000 - 400,000,000 \\
        Gross (USD): \\
            & t5 - Below median: 106,457 - 347,280,359\\
            & t6 - Above median: 347,280,359 - 2,796,000,000\\
        Release date: \\
            & t7 - Late 1980's and 1990's \\
            & t8 - Early 2000's \\
            & t9 - 2010's and 2020's \\
        Genre: \\
            & t10 - Feel good or drama \\
            & t11 - Action/adventure or thriller \\
            & t12 - Family/sci-fi/fantasy \\
            & t13 - Real world adaptation \\
        Runtime (minutes): \\
            & t14 - Below median: 79 - 117 \\
            & t15 - Above median: 117 - 201\\
        Rating: \\
            & t16 - Below median: 4.1 - 6.9\\
            & t17 - Above median: 6.9 - 9.0\\
        Rating-count (number of ratings): \\
            & t18 - Below median: 242 - 251,771 \\
            & t19 - Above median:  251,771 - 2,127,228 \\
    \bottomrule
  \end{tabular}
    \caption{Tag Definitions for Movies}
    \label{tab:movie_tags}
\end{table}

\begin{figure}
\centering
\begin{minipage}{.47\textwidth}
  \hspace*{-0.1in}
  \includegraphics[width=7cm]{{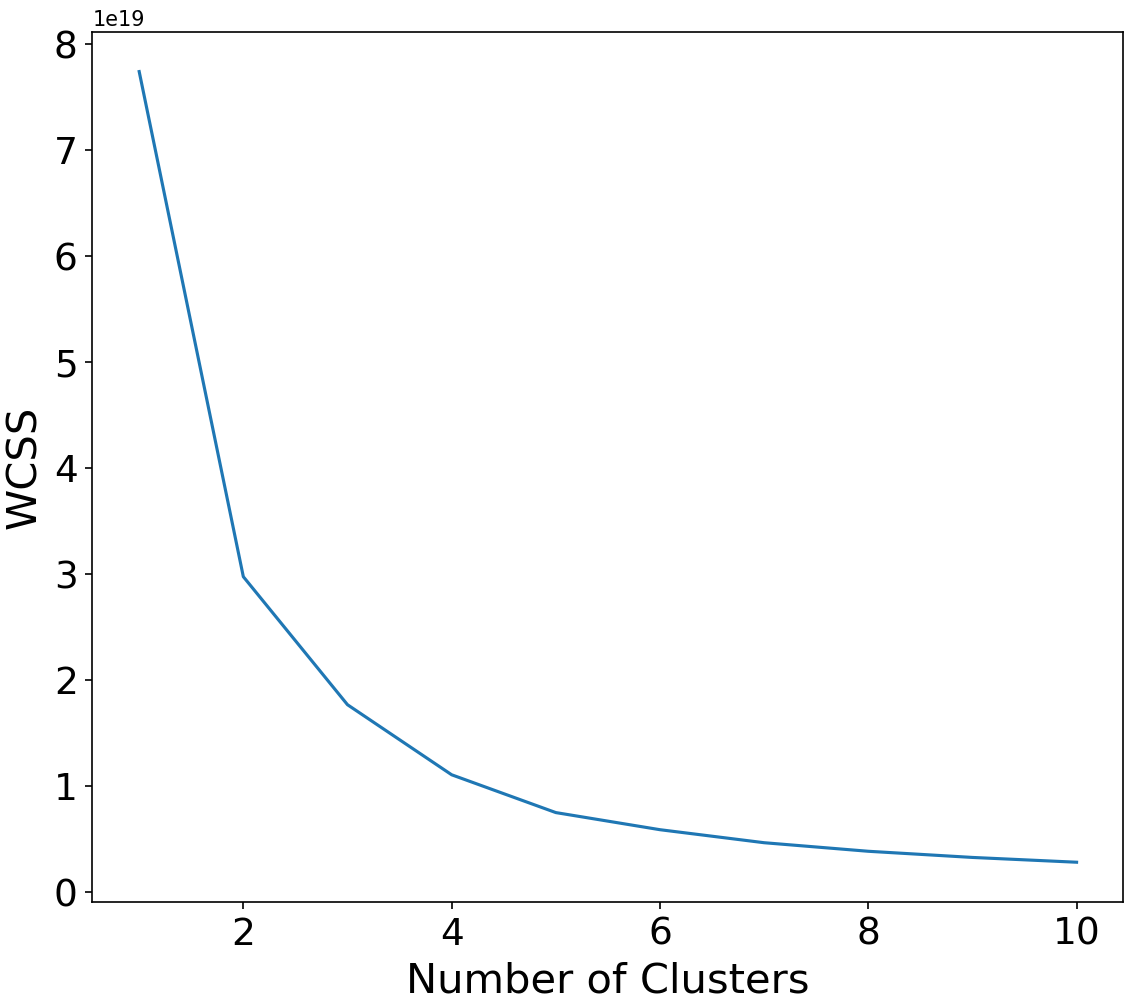}}
  \captionof{figure}{Movies Data: 
  Elbow Method.}
  \label{fig:movies_elbow_method}
\end{minipage}%
\quad 
\begin{minipage}{.47\textwidth}
  \includegraphics[width=7cm]{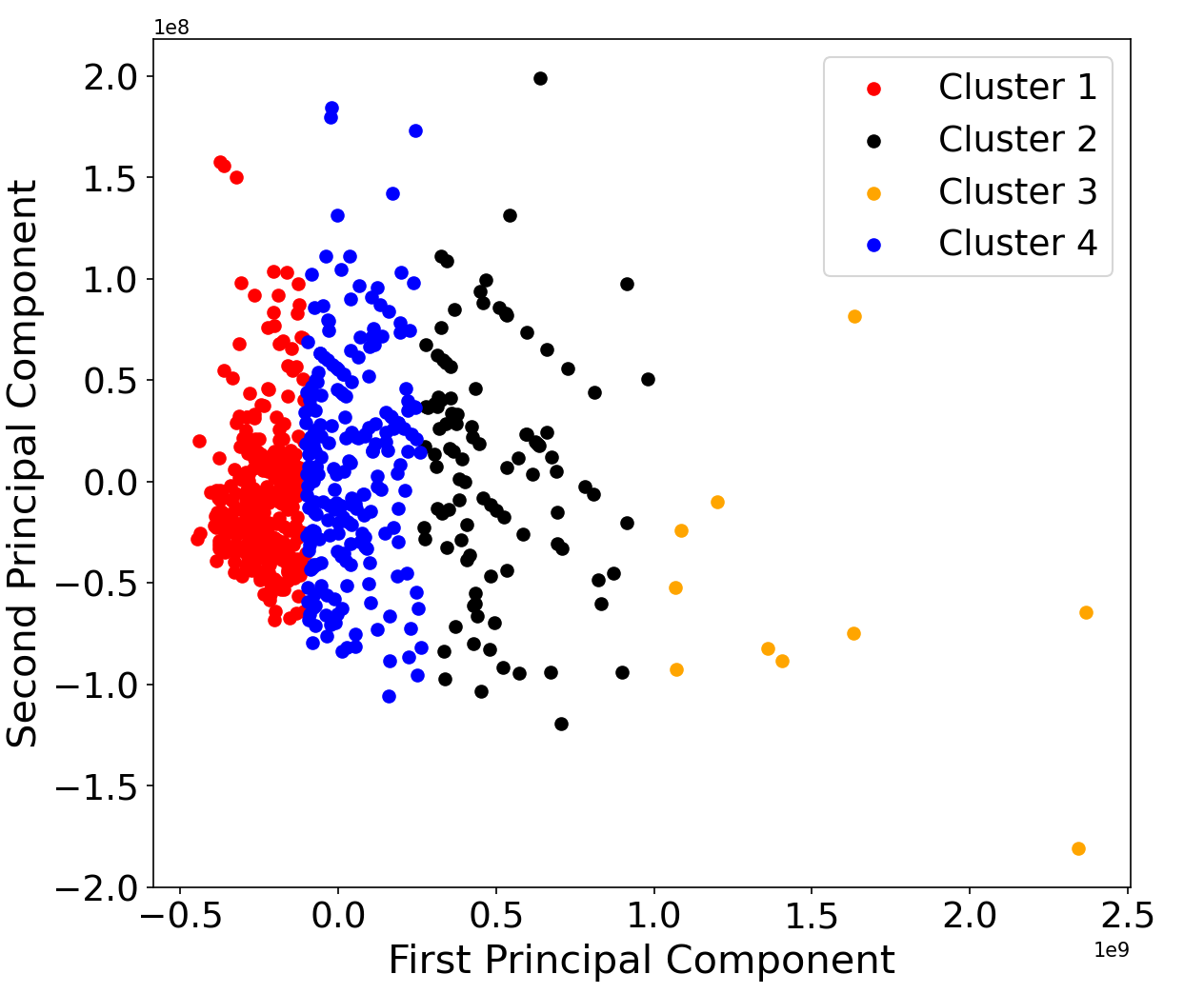}
  \captionof{figure}{Movies Data: Principal Component Analysis.}
  \label{fig:movies_pca}
\end{minipage}
\end{figure}

Using the elbow method, which evaluates the sum of squared errors for an increasing number of clusters, 4 clusters were determined to be the optimal number (see Figure \ref{fig:movies_elbow_method}).

{\footnotesize
\begin{table}[th!]
\centering
\begin{tabular}{|>{\centering\arraybackslash}m{2.8cm}|>{\centering\arraybackslash}m{2.8cm}|>{\centering\arraybackslash}m{2.8cm}|>{\centering\arraybackslash}m{2.8cm}|>{\centering\arraybackslash}m{2.8cm}|}
\hline
\textbf{} & {\textbf{Cluster 1:}\newline \textbf{104 movies}} & {\textbf{Cluster 2:}\newline \textbf{315 movies}} & {\textbf{Cluster 3:}\newline \textbf{11 movies}} & {\textbf{Cluster 4:}\newline \textbf{208 movies}} \\ \hline\hline
\textbf{Disjunctive Heuristic} & [t6] & [t5, t3] & [t4] & [t6, t4] \\ \hline
\textbf{Disjunctive Exact} & [t6] & [t16, t17] & [t6] & [t16, t17] \\ \hline
\textbf{CNF} & ([t6], [t4, t15, t9, t17]) & ([t5, t3], [t18, t2, t14, t7]) & ([t4], [t6]) & ([t6, t4], [t2, t17, t15, t14]) \\ \hline
\end{tabular}
\caption{Movies: Hitting Sets}
\label{table:movies_hitting_sets}
\end{table}
}

Further information concerning the ``genres" tags can be found in Appendix B. The tag definitions were applied to the clusters using the ILP formulation (Section~\ref{sse:ilp_for_mhs}), resulting in the findings illustrated in Table \ref{table:movies_hitting_sets}. These results can be interpreted with the following generalizations:\\ \\
\textbf{Cluster 1}: Movies that generated significant revenue. \\
\textbf{Cluster 2}: Low-budget movies with low revenue. \\
\textbf{Cluster 3}: High-budget movies. \\
\textbf{Cluster 4}: High-budget movies with significant revenue. \\

Some of the hitting sets contain the same tags (e.g., Cluster 3 and Cluster 4 both contain t4); this issue arises when optimizing for minimal cluster size, resulting in less unique hitting sets. To address this, a helper algorithm was developed to filter two clusters against each other. If a tag appears in more than a user-specified percentage for both clusters, it will not be considered for the disjunctive hitting set generation of those two clusters. This method aims to provide more information about the clusters by restricting the use of tags that occur prevalently, necessitating the utilization of other, more unique tags for hitting set generation.

Using this filtering method, with a maximum tolerated shared percentage of 50\% for Clusters 1 and 2, and Clusters 3 and 4, new disjunctive heuristic hitting sets were generated. Essentially, if a tag occurs in more than 50\% of the data for Cluster 1 and Cluster 2, it will not be considered for hitting set generation for those clusters. The same applies to Clusters 3 and 4.

The hitting sets produced by this filter are shown in Table \ref{tab:movies_results_filtered}, and additional information for Clusters 3 and 4 can be interpreted as shown in Table \ref{tab:cluster_filtered_interpretation}.

{\footnotesize
\begin{table}[h!]
    \centering
    \begin{tabular}{| m{3cm} | m{3.5cm} | m{7cm} |}
        \hline
        \textbf{Clusters} & \textbf{Percentage Threshold} & \textbf{Filtered Disjunctive Heuristic} \\ \hline\hline
        Cluster 1 & At most 56\% & \texttt{[t6]} \\ \hline
        Cluster 2 & At most 56\% & \texttt{[t5, t3]} \\ \hline
        Cluster 3 & At most 50\% & \texttt{[t9, t11, t10]} \\ \hline
        Cluster 4 & At most 50\% & \texttt{[t14, t9, t8, t12, t11, t1, t3]} \\ \hline
    \end{tabular}
    \caption{Filtered Disjunctive Heuristics for Clusters}
    \label{tab:movies_results_filtered}
\end{table}
}

\begin{table}[htbp]
    \centering
    \begin{adjustbox}{max width=\textwidth}
    \begin{tabular}{@{}p{0.4\linewidth}p{0.6\linewidth}@{}}
        \toprule
        \textbf{Cluster} & \textbf{Characteristics} \\
        \midrule
        Cluster 3 & Recently released movies with ``feel-good" / drama or action / adventure / thriller genres \\
        Cluster 4 & Short runtime / released in the range 2000-present / family, sci-fi, or fantasy / action, adventure, or thriller / rated G or PG / low budget \\
        \bottomrule
    \end{tabular}
    \end{adjustbox}
    \caption{Interpretation of Cluster 3 and 4 Results Post-Filter}
    \label{tab:cluster_filtered_interpretation}
\end{table}

The contents of the four clusters generated, as well as principal component visualizations of the tag distributions within the four clusters, can be found in Appendix~B.

The application of the disjunctive and CNF hitting set solver models to this movies dataset demonstrates some practical applications. These hitting sets provide interesting insights.  Cluster~1, containing movies that generated significant profit, offers information regarding the types of movies that generally experience great financial success regardless of budget. Cluster 2 serves as a cautionary guide for stakeholders, highlighting low-budget movies that failed to generate substantial revenue and providing insights for budget allocation and risk assessment in low-budget ventures. Cluster 3 showcases high-budget movies, which are of interest to studios aiming to understand the intricacies of resource-intensive projects. Finally, Cluster 4 illustrates high-budget movies that achieved financial success, serving as a blueprint for studio investors seeking to maximize returns on large investments.

Overall, the ability of the disjunctive and CNF hitting set solvers to classify movies based on financial performance can inform decision-making processes within the film industry. This experiment also underscores the practicality of this cluster explanation method for datasets with more than two clusters.

\subsection{Divorce Predictors}
\begin{figure}
\centering
\begin{minipage}{.47\textwidth}
  \hspace*{-0.2in}
  \includegraphics[width=7cm]{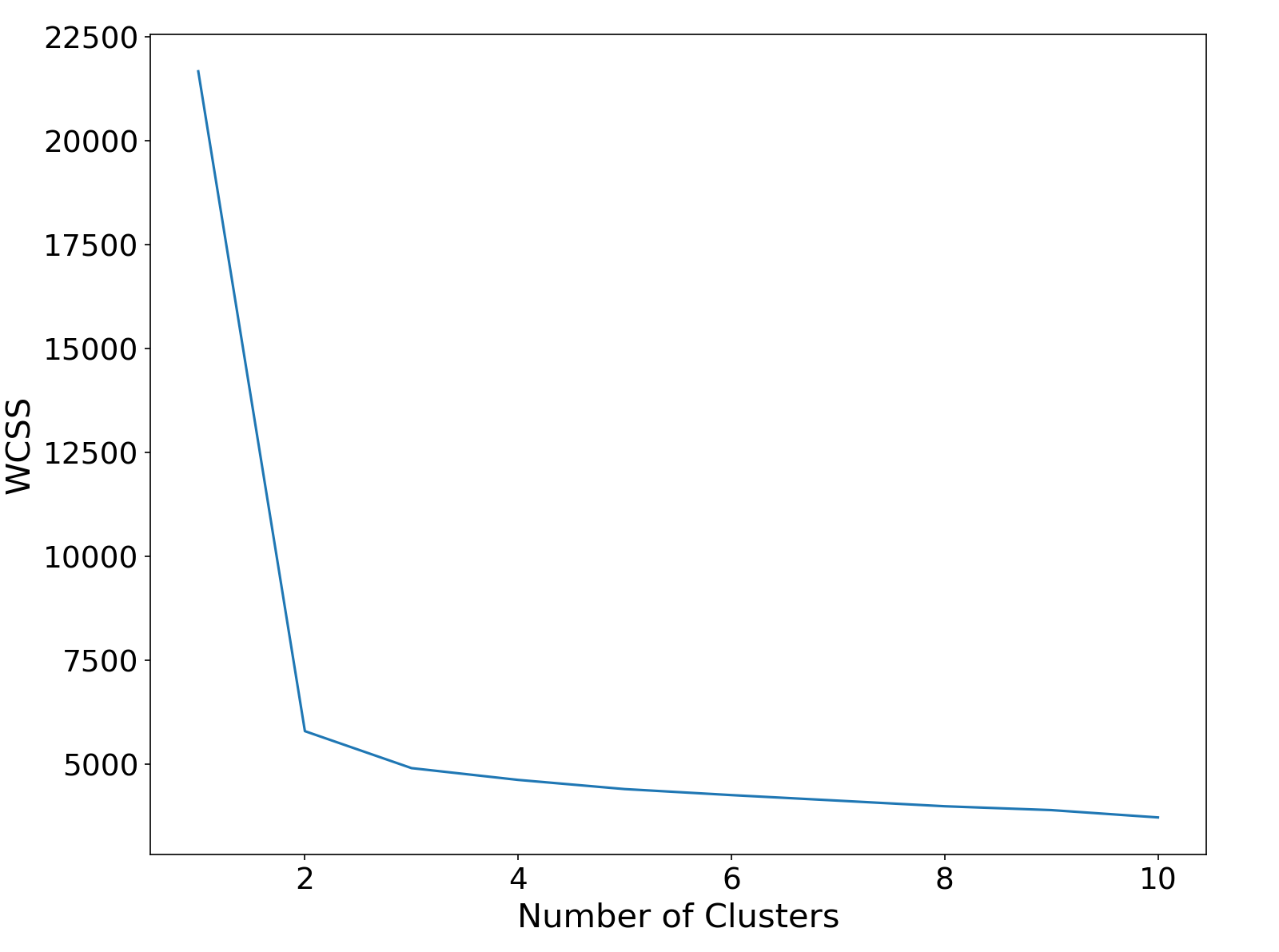}
  \captionof{figure}{Divorce Predictors Data:\newline 
  Elbow Method.}
  \label{fig:divorce_elbow_method}
\end{minipage}%
\quad 
\begin{minipage}{.47\textwidth}
  \includegraphics[width=7cm]{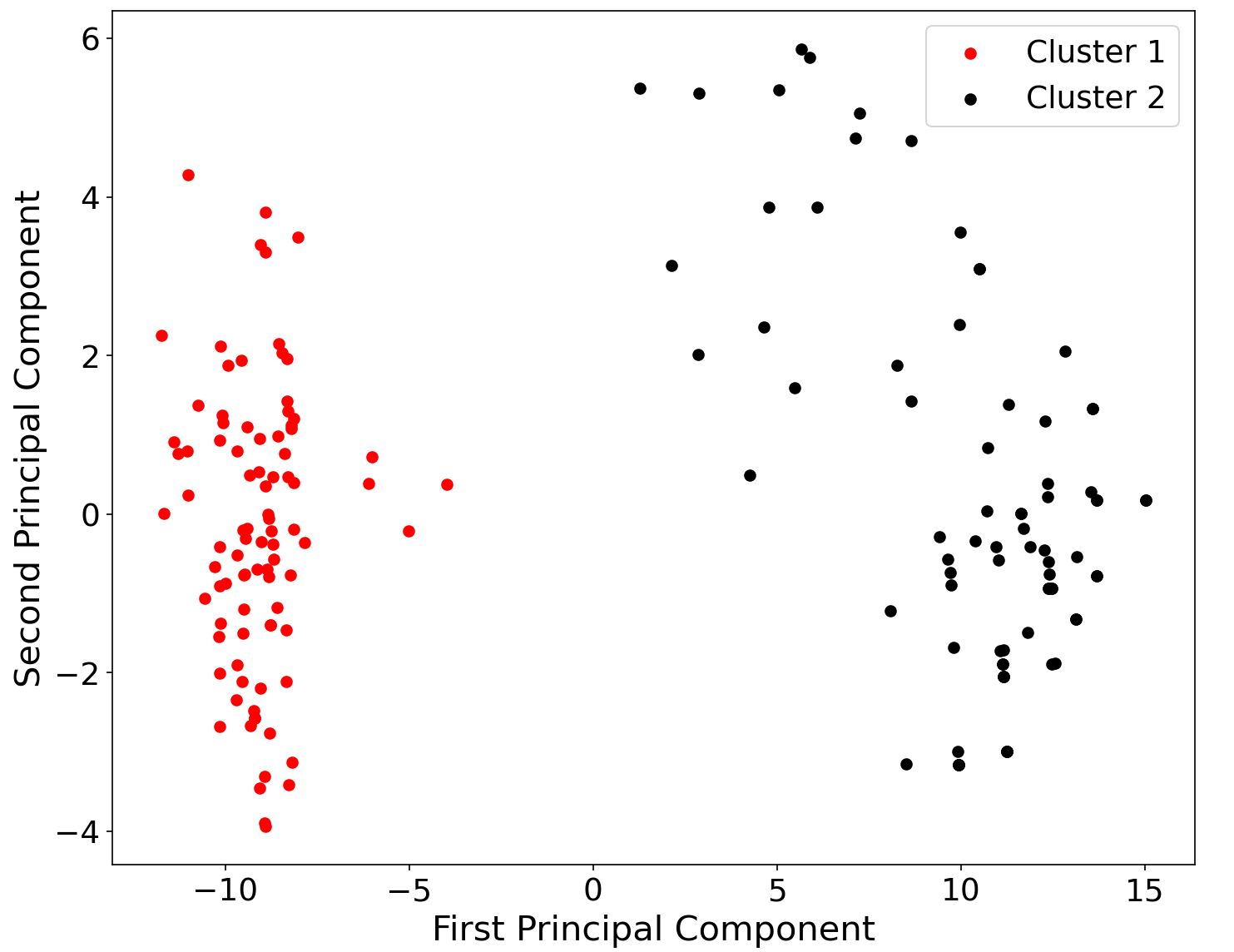}
  \captionof{figure}{Divorce Predictors Data: Principal Component Analysis.}
  \label{fig:divorce_pca}
\end{minipage}
\end{figure}

The goal of this experiment is to cluster married couples based on their responses to a divorce predictors survey, utilizing tags to identify commonalities among couples within the same cluster. This information is used to infer possible reasons behind their cluster membership. The general process involves determining the appropriate number of clusters, clustering the data, and defining tags that can be interpreted by hitting set solver algorithms. Using the elbow method, the optimal number of clusters for this dataset was determined to be 2 clusters (see Figure \ref{fig:divorce_elbow_method}). Figure \ref{fig:divorce_pca} displays the results of the principal component analysis for these two clusters.

The couples were divided somewhat evenly by $k$-means with 90 couples in the
first cluster and 80 couples in the second.

The tagging method that produced distinct hitting sets for the two clusters was
found by defining tags based on whether couples had a “high” or “low” level of
agreement with each statement in the survey. If a couple gave a rating of a 0-2, this
would be classified as a “low” level of agreement whereas a rating of 3-4 would be
classified as a “high” level of agreement. Thus, if a couple has tag t1, this would
mean that the couple does not agree highly with statement No.~1 of the survey,
whereas tag t2 would mean that they do agree strongly with the first statement. Tag
t3 would indicate that the couple does not agree strongly with statement No.~2
whereas t4 would mean that they do agree 
strongly, etc.

These tags were then applied to the data that forms the input to the four hitting set solver algorithms (disjunctive heuristic, CNF heuristic, DNF
heuristic, and disjunctive exact) to obtain descriptors for each of the two clusters.

Since not much information is available about the content of the clusters beyond the hitting
sets, further cluster explanation is provided whereby the percentage of
the data in each cluster which has any given tag is calculated. These
percentages indicate what percent of the data in a cluster had any given tag. For
instance, we know from the disjunctive heuristic and exact models that every couple in
Cluster 1 has the tag values t5 and t7; therefore, we would expect that the percentages
for t5 and t7 in cluster 1 would be 100 percent, which is indeed the case. These percentages can be found in Appendix~B (see Table \ref{tab:divorce_tag_percentages}).

Cluster~1, containing 90 couples, has its disjunctive heuristic hitting set containing tags t5 and t7, which correspond to low agreement with the statements ``When we need it, we can take our discussions from the beginning and correct it" and ``When I discuss with my spouse, contacting them will eventually work," respectively. Cluster~2, containing 80 couples, has a disjunctive heuristic hitting set with tags t20 and t38, indicating high agreement with ``Most of our goals are common" and ``We have similar ideas about marriage roles." The disjunctive exact hitting set for Cluster~2 includes tags t34 and t53, indicating high agreement with ``We share the same views about being happy in life" and low agreement with ``I know what my spouse's current sources of stress are." For both clusters, the CNF form of the explanation provides no additional information beyond the tags in the disjunctive sets (see Table \ref{table:divorce_hitting_sets}).

\begin{table}[h!]
\centering
\begin{tabular}{|>{\centering\arraybackslash}m{3cm}|>{\centering\arraybackslash}m{5cm}|>{\centering\arraybackslash}m{5cm}|}
\hline
\textbf{} & \textbf{Cluster 1: 90 couples} & \textbf{Cluster 2: 80 couples} \\ \hline
\textbf{Disjunctive Heuristic} & [t5] & [t20, t38] \\ \hline
\textbf{Disjunctive Exact} & [t7] & [t34, t53] \\ \hline
\textbf{CNF} & ([t5], [t7]) & ([t20, t38], [t18, t40]) \\ \hline
\end{tabular}
\caption{Divorce Predictors: Hitting Sets}
\label{table:divorce_hitting_sets}
\end{table}

Interpreting these results, Cluster~1 suggests potential communication challenges or difficulties in resolving conflicts. Couples in this cluster may struggle with effectively communicating their needs and resolving disagreements.
In Cluster~2, although couples agree on common goals and marriage roles, they exhibit low agreement on understanding each other's sources of stress. This suggests that while they may communicate well about shared objectives, there may be a lack of depth in their communication regarding personal concerns and stress.

These findings could provide useful insights for relationship counselors or therapists. Interventions for couples with survey responses similar to those in Cluster~1 could focus on improving communication skills and conflict resolution strategies. For couples in Cluster~2, interventions could aim to deepen emotional connections and advocate for a better understanding of each other's individual needs and stressors.


\section{Summary, Limitations and Future Work}
\label{sec:concl}

We applied the tag-based approach from~\cite{DGR-NIPS-2018} for
developing post-hoc explanations of the outputs of clustering
algorithms. Specifically, we considered methods to construct
disjunctive descriptors and CNF-style descriptors with two clauses.
We applied these methods to several datasets and outlined some insights
resulting from the explanations.  We also examined the scalability
of our methods using synthetic datasets.

One limitation of our work is that it relies on the availability
of tags for each data item.  For small datasets, it may be possible
to construct  tags manually.  Commercial software such as Google
Photos can generate tags for images with people.  Some tag editors
for images in objects are also available~\cite{booru-2022}.  When
the features are interpretable, one can use feature values (or
suitable ranges) to construct appropriate tags. Our work used this
approach for selecting tags.  A second limitation of our work is
that for CNF-style descriptors, one needs to use suitable data-dependent
methods (such as filters used in our work) when it is difficult to
get two disjoint descriptors.  In practice, one may overcome this
limitation by not requiring the descriptors to cover all the data
items in a cluster as done in~\cite{Sambaturu-etal-2020}.

We close by pointing out some directions for future work. 
One direction is to cluster a dataset using different clustering algorithms, develop descriptors
for the solutions and study which of the clustering
methods lead to more explainable solutions.
A second direction is to make the notion of a descriptor stronger in the following manner.
Our work requires a descriptor for a cluster to contain just one tag describing each data item in the cluster. It will be useful to study schemes where the coverage requirement is stronger; that is, one may require a descriptor where each data item
must have at least $\ell{} \geq 2$ tags in the
descriptor.
An even more general version is to  
consider a scheme where each tag has a certain significance level. In such a version of the problem, the goal will be to find a descriptor
such that the sum of the significance levels of
the tags that describe a data item is at least a given parameter $\alpha$.
Finally, it will also be useful to develop tag-based explanation methods for solutions produced 
by non-partitional clustering algorithms.
 

\bigskip

\bigskip 

\noindent
\textbf{Acknowledgment:}
This work was partially supported by NSF Grant
IIS-1908530 titled ``Explaining Unsupervised Learning: Combinatorial
Optimization Formulations, Methods and Applications''.

 \bigskip

\bibliographystyle{plain}
\bibliography{references}

\clearpage

\section{Appendix A: Results for the Census Dataset}
In this appendix, we present our experimental results for an additional 
dataset, namely Census.
A description of this dataset was provided in
Section~\ref{sec:data_desc}.

\begin{figure}
\centering
\begin{minipage}{.47\textwidth}
  \hspace*{-0.1in}
  \includegraphics[width=7cm]{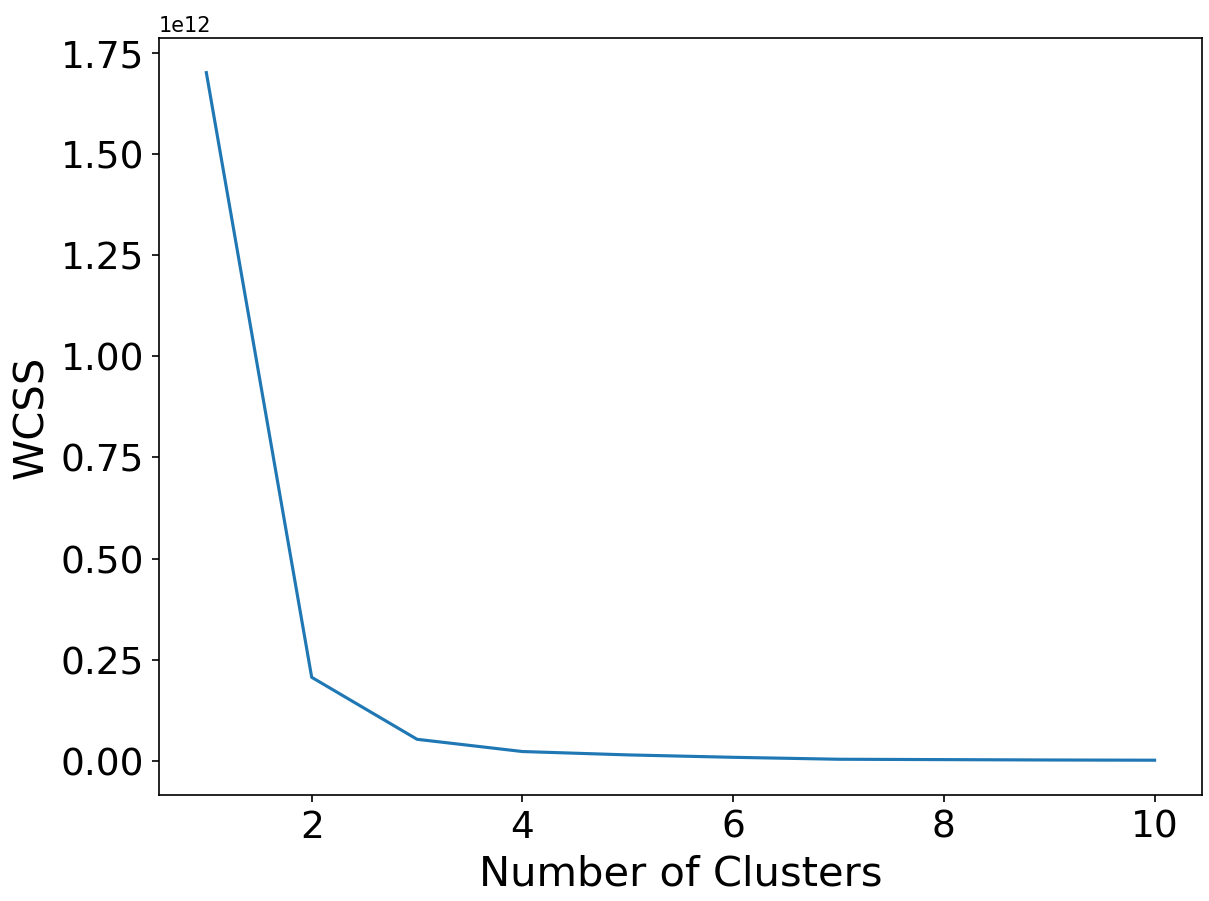}
  \captionof{figure}{Census Data: 
  Elbow Method.}
  \label{fig:census_elbow_method}
\end{minipage}%
\quad 
\begin{minipage}{.47\textwidth}
  \includegraphics[width=7cm]{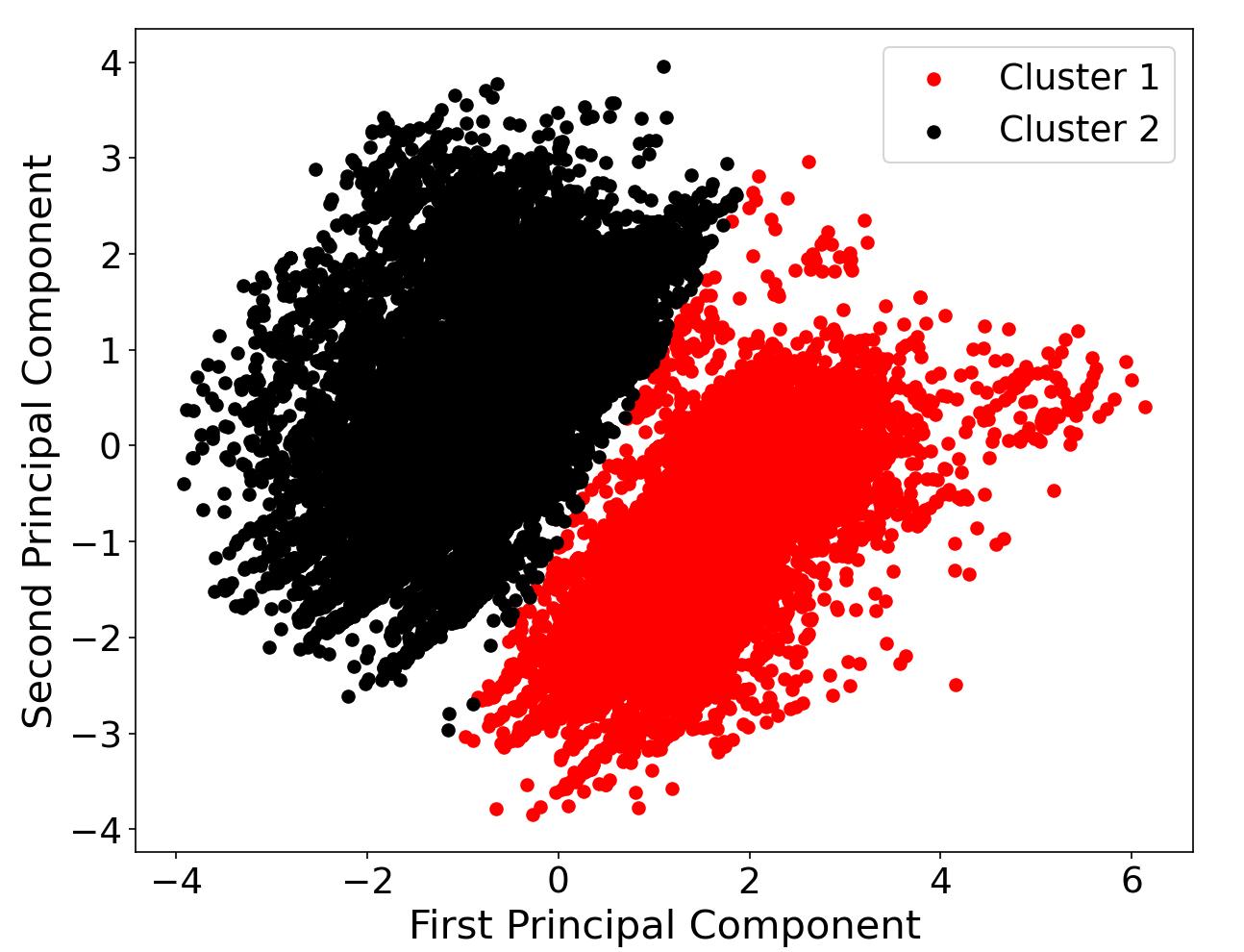}
  \captionof{figure}{Census Data: Principal Component Analysis.}
  \label{fig:census_pca}
\end{minipage}
\end{figure}

The $k$-means algorithm was on the data to produce two clusters,
yielding disparate cluster sizes. The first cluster notably outnumbered
the second one, comprising 25,518 individuals compared to only 5,201
in the second cluster.

To generate distinct hitting sets for the two clusters, a tagging
approach was employed. This involved calculating the median or mean
values for continuous data and assigning dummy integer labels for
categorical columns. For instance, age was categorized into two
tags based on whether an individual was younger or older than the
median age of 37 years. Similarly, each race was assigned a unique
tag. The resulting tag definitions are illustrated in
Table~\ref{census_tag_definitions}.

\smallskip 

\renewcommand{\arraystretch}{1.2}
\setlength{\tabcolsep}{4pt}

{\small 
\begin{longtable}{|>{\centering\arraybackslash}m{0.15\textwidth}|>{\centering\arraybackslash}m{0.27\textwidth}|>{\centering\arraybackslash}m{0.35\textwidth}|}
\hline
\textbf{Tags} & \textbf{Attribute} & \textbf{Definition Method} \\ \hline\hline
\endfirsthead
\hline
\textbf{Tags} & \textbf{Attribute} & \textbf{Definition Method} \\ \hline
\endhead
t1 / t2 & Age less / greater 37 years & Use median \\ \hline
t3 / t4 / t5 & No income / govt. employee / private or other & Use categorical info in dataframe \\ \hline
t6 / t7 & Does not have / has at least bachelors degree & Use categorical info in dataframe \\ \hline
t8 / t9 & Number of years of education less / greater than 10 years & Use median \\ \hline
t10 / t11 & Married / not married & Use categorical info in dataframe \\ \hline
t12 / t13 / t14 & General Service / Corporate Support / Specialist or Managerial position & Use categorical info in dataframe \\ \hline
t15 / t16 / t17 / t18 / t19 & White / Asian-Pac-Islander / Amer-Indian-Eskimo / Black / Other & Use categorical info in dataframe \\ \hline
t20 / t21 & Female / Male & Use categorical info in dataframe \\ \hline
t22 / t23 & Capital gain below / above \$1086 & Use mean since most individuals have a gain of 0 \\ \hline
t24 / t25 & Capital loss below / above \$87 & Use mean since most individuals have a loss of 0 \\ \hline
t26 / t27 & Hours per week below / above 40 hours per week & Use median \\ \hline
t28 / t29 & Developing / developed country & Use categorical info in dataframe \\ \hline
\caption{Census Data Tags}
\label{census_tag_definitions}
\end{longtable}
}

These tags were subsequently used by our three hitting set algorithms: disjunctive heuristic, CNF heuristic, and disjunctive exact. The outcomes of these algorithms are depicted in Table~\ref{table:census_hitting_sets}.

\begin{table}[h!]
\centering
\begin{tabular}{|>{\centering\arraybackslash}m{3cm}|>{\centering\arraybackslash}m{6cm}|>{\centering\arraybackslash}m{6cm}|}
\hline
\textbf{} & \textbf{Cluster 1: 5201 individuals} & \textbf{Cluster 2: 25518 individuals} \\ \hline
\textbf{Disjunctive Heuristic} & [t18] & [t19, t3, t11, t20, t17, t4, t7] \\ \hline
\textbf{Disjunctive Exact} & [t18] & [t20, t21] \\ \hline
\textbf{CNF} & ([t18], [t20, t3, t11, t4, t16, t7]) & ([t19, t3, t11, t20, t17, t4, t7], [t6, t10, t5, t9, t16, t8]) \\ \hline
\end{tabular}
\caption{Census: Hitting Sets}
\label{table:census_hitting_sets}
\end{table}

As with the datasets, the application of hitting set solvers to the tagsets for each cluster yields insightful findings that contribute to a better understanding of the cluster contents.

The disjunctive heuristic algorithm produced a hitting set for Cluster~1 defined by tag t18, corresponding to the definition of ``black ethnicity." Additionally, the conjunctive normal form explanation (CNF) provided further insights into Cluster~1, emphasizing the most relevant tags that span the entire cluster, in the absence of tag t18 during hitting set generation. Notably, there are more tags in the second clause of the CNF than the first, indicating the necessity for additional tags to cover all the data items in the cluster when the most pertinent tag is absent.

For Cluster~2, the disjunctive heuristic found that all data could be explained by tags t20 and t21, corresponding to ``female" and ``male," respectively. However, stating that the entire cluster consists of females and males alone does not offer useful information about the cluster's characteristics. 
We used the method discussed in 
Section~\ref{sse:expt_majors} to prevent the generation of hitting sets consisting solely of complementary tags.

Table~\ref{tab:census_clusters_percentages} displays the hitting sets generated by the disjunctive heuristic, CNF heuristic, and disjunctive exact algorithms for both clusters, along with the percentage of data containing each tag within each respective cluster.

\renewcommand{\arraystretch}{1.2}
\setlength{\tabcolsep}{4pt}

\begin{table}[tbhp]
\centering
\caption{Meaning of Hitting Sets for Clusters with Tag Percentages}
\label{tab:census_clusters_percentages}
\begin{adjustbox}{max width=\textwidth}
\begin{tabular}{|p{6cm}|p{9cm}|}
\hline
\multicolumn{2}{|c|}{\textbf{Cluster 1 (5201 individuals)}} \\ \hline
Approximate (Heuristic) Hitting Set & [Black (100\%)] \\ \hline
CNF Descriptor & (Black (100\%), [Female (93.308\%), No income (90.846\%), Not married (84.808\%), Govt. employee (79.135\%), Asian-Pac-Islander (52.288\%), has at least bachelors degree (50.692\%)]) \\ \hline
Minimum (Exact) Hitting Set & [Black (100\%)] \\ \hline
\multicolumn{2}{|c|}{} \\ \hline
\multicolumn{2}{|c|}{\textbf{Cluster 2 (25,518 individuals)}} \\ \hline
Approximate (Heuristic) Hitting Set & [``Other" ethnicity (91.723\%), No income (85.7\%), Not married (85.621\%), Female (72.516\%), Amer-Indian-Eskimo (70.827\%), Govt. employee (65.498\%), Has at least bachelors degree (55.398\%)] \\ \hline
CNF Descriptor & ([``Other" ethnicity (91.723\%), No income (85.7\%), Not married (85.621\%), Female (72.516\%), Amer-Indian-Eskimo (70.827\%), Govt. employee (65.498\%), Has at least bachelors degree (55.398\%)], [Does not have at least bachelors degree (44.602\%), Married (39.048\%), Privately employed/non-govt position (34.502\%), Number of years of education less than 10 years (28.883\%), Asian-Pac-Islander (29.173\%), Number of years of education less than 10 years (28.883\%)]) \\ \hline
Minimum (Exact) Hitting Set & [Female (72.516\%), Male (27.484\%)] \\ \hline
\end{tabular}
\end{adjustbox}
\end{table}

From the hitting sets generated for each cluster, clear differences in content can be observed. Cluster~1, the smaller of the two clusters, consists entirely of individuals identifying as ethnically black, making black ethnicity the primary defining characteristic. The CNF descriptor further clarifies the characteristics of Cluster~1. In addition to ethnicity, a large percentage of individuals are female, unmarried, have low to no income, and a smaller but still significant portion have some form of government employment (which may include military or protective services), with approximately half holding a bachelor's degree. Cluster~1 represents a subgroup predominantly composed of Black individuals, with a notable prevalence of females (93.308\%) and individuals with no stable income (90.846\%). The majority of these individuals are unmarried (84.808\%), suggesting significant socio-economic challenges within this demographic group.

In contrast, Cluster~2 appears to represent a more diverse subgroup, consistent with its larger size compared to Cluster~1. Like Cluster~1, Cluster~2 contains a significant portion of individuals with no income (85.7\%), females (72.516\%), and unmarried individuals (85.621\%), highlighting a sampling bias in the dataset. Cluster~2 also includes additional information that distinguishes it from Cluster~1. The prime defining characteristic of Cluster 2 appears to be the ``other" ethnicity tag, suggesting that the k-means clustering algorithm identified ethnicity as a discriminative feature in the dataset.

Understanding these clusters and their distinctive traits can be beneficial for crafting targeted interventions, shaping policy decisions, and refining marketing strategies tailored to specific socio-economic needs. The disjunctive and CNF cluster explanation methods can aid in quickly assessing the defining characteristics of clusters, thus highlighting areas for intervention and further analysis.

\section{Appendix B}
Here, we provide some of the outputs generated
by our experiments on various datasets.
These outputs were used in generating the
explanations.
We decided to include these in an appendix
since we wanted the main paper's focus to be
on the explanations generated from the descriptors 
produced by our methods. A brief description
of the outputs included in this appendix is provided below.

\begin{itemize}
\item For the Divorce Predictors dataset, 
Table~\ref{tab:divorce_tag_percentages} gives
the percentage of data items covered by each tag
in the two clusters.

\item For the Movies dataset,
the actual genres defined by the dataset are listed in Table~\ref{tab:movie_categories}.
The clusters in which various movies were placed 
are shown in Tables~\ref{tab:movies_one_pone} through \ref{tab:movies_four_ptwo}.
Table~\ref{tab:movies_tag_percentages} gives
the percentage of data items covered by each tag
in the four clusters.

\item For the Census dataset, 
Table~\ref{tab:census_original_features} gives the features of data items appearing in the dataset and 
Table~\ref{tab:census_tag_percentages} gives
the percentage of data items covered by each tag
in the two clusters.

\item For the College Majors dataset, 
Tables~\ref{tab:majors_one_pone} through 
\ref{tab:majors_three} give the fields of study appearing in the three clusters.
Table~\ref{tab:majors_tag_percentages} gives
the percentage of data items covered by each tag
in the three clusters.
\end{itemize}

\begin{table}[h!]
    \centering
    \small 
    \setlength{\tabcolsep}{5pt} 
    \renewcommand{\arraystretch}{1.2} 
    \begin{tabular}{| m{1.3cm} | m{1.8cm} | m{1.8cm} | m{1.8cm} | m{1.8cm} | m{1.8cm} | m{1.8cm} | m{1.8cm} | m{1.8cm} | m{1.8cm} | m{1.8cm} | m{1.8cm} | m{1.8cm} | m{1.8cm} | m{1.8cm} |}
        \hline \textbf{Cluster}
        & \textbf{t1-t18} & \textbf{t19-t36} & \textbf{t37-t54} & \textbf{t55-t72} & \textbf{t73-t90} & \textbf{t91-t108}
        \\ \hline
        1 & t1=95.556 t2=4.444 t3=94.444 t4=5.556 t5=100.0 t6=0.0 t7=100.0 t8=0.0 t9=100.0 t10=0.0 t11=98.889 t12=1.111 t13=98.889 t14=1.111 t15=100.0 t16=0.0 t17=100.0 t18=0.0 & t19=100.0 t20=0.0 t21=100.0 t22=0.0 t23=100.0 t24=0.0 t25=100.0 t26=0.0 t27=100.0 t28=0.0 t29=100.0 t30=0.0 t31=95.556 t32=4.444 t33=97.778 t34=2.222 t35=98.889 t36=1.111 & t37=96.667 t38=3.333 t39=97.778 t40=2.222 t41=98.889 t42=1.111 t43=67.778 t44=32.222 t45=70.0 t46=30.0 t47=83.333 t48=16.667 t49=81.111 t50=18.889 t51=86.667 t52=13.333 t53=85.556 t54=14.444 & t55=0.0 t56=0.0 t57=0.0 t58=0.0 t59=0.0 t60=0.0 t61=0.0 t62=0.0 t63=0.0 t64=0.0 t65=0.0 t66=0.0 t67=0.0 t68=0.0 t69=0.0 t70=0.0 t71=0.0 t72=0.0 & t73=0.0 t74=0.0 t75=0.0 t76=0.0 t77=0.0 t78=0.0 t79=0.0 t80=0.0 t81=0.0 t82=0.0 t83=0.0 t84=0.0 t85=0.0 t86=0.0 t87=0.0 t88=0.0 t89=0.0 t90=0.0 & t91=0.0 t92=0.0 t93=0.0 t94=0.0 t95=0.0 t96=0.0 t97=0.0 t98=0.0 t99=0.0 t100=0.0 t101=0.0 t102=0.0 t103=0.0 t104=0.0 t105=0.0 t106=0.0 t107=0.0 t108=0.0 \\ \hline
        2 & t1=7.5 t2=92.5 t3=20.0 t4=80.0 t5=12.5 t6=87.5 t7=88.75 t8=11.25 t9=17.5 t10=82.5 t11=6.25 t12=93.75 t13=11.25 t14=88.75 t15=15.0 t16=85.0 t17=5.0 t18=95.0 & t19=3.75 t20=96.25 t21=27.5 t22=72.5 t23=18.75 t24=81.25 t25=16.25 t26=83.75 t27=27.5 t28=72.5 t29=17.5 t30=82.5 t31=8.75 t32=91.25 t33=11.25 t34=88.75 t35=12.5 t36=87.5 & t37=3.75 t38=96.25 t39=5.0 t40=95.0 t41=7.5 t42=92.5 t43=8.75 t44=91.25 t45=12.5 t46=87.5 t47=15.0 t48=85.0 t49=11.25 t50=88.75 t51=7.5 t52=92.5 t53=13.75 t54=86.25 & t55=0.0 t56=0.0 t57=0.0 t58=0.0 t59=0.0 t60=0.0 t61=0.0 t62=0.0 t63=0.0 t64=0.0 t65=0.0 t66=0.0 t67=0.0 t68=0.0 t69=0.0 t70=0.0 t71=0.0 t72=0.0 & t73=0.0 t74=0.0 t75=0.0 t76=0.0 t77=0.0 t78=0.0 t79=0.0 t80=0.0 t81=0.0 t82=0.0 t83=0.0 t84=0.0 t85=0.0 t86=0.0 t87=0.0 t88=0.0 t89=0.0 t90=0.0 & t91=0.0 t92=0.0 t93=0.0 t94=0.0 t95=0.0 t96=0.0 t97=0.0 t98=0.0 t99=0.0 t100=0.0 t101=0.0 t102=0.0 t103=0.0 t104=0.0 t105=0.0 t106=0.0 t107=0.0 t108=0.0 \\ \hline
    \end{tabular}
    \caption{Divorce Predictors: Tag Percentages}
    \label{tab:divorce_tag_percentages}
\end{table}


\begin{table}[htbp]
  \centering
  \caption{Movie Categories}
  \label{tab:movie_categories}
  \begin{tabular}{ll}
    \toprule
    \textbf{Category} & \textbf{Movies} \\
    \midrule
    \multirow{6}{*}{Feel Good or Drama} & Romantic Comedy \\
                                         & Comedy, Romance, Music \\
                                         & Musical \\
                                         & Drama \\
                                         & Romance \\
                                         & Comedy \\
    \midrule
    \multirow{5}{*}{Action/Adventure or Thriller} & Adventure \\
                                                   & Thriller \\
                                                   & Action \\
                                                   & Horror \\
                                                   & Western \\
    \midrule
    \multirow{4}{*}{Family/Sci-Fi/Fantasy} & Family \\
                                             & Animation \\
                                             & Science Fiction \\
                                             & Fantasy \\
    \midrule
    \multirow{4}{*}{Real World Adaptation} & Crime \\
                                            & War \\
                                            & Mystery \\
                                            & History \\
    \bottomrule
  \end{tabular}
\end{table}

\begin{table}[ht]
    \centering
    \small
    \begin{tabular}{|p{1\textwidth}|}
    \hline
    \textbf{Cluster 1 Movies (Part 1)} \\
    \hline
    The Lion King, Independence Day, Star Wars: Episode I - The Phantom Menace, The Lord of the Rings: The Fellowship of the Ring, Harry Potter and the Sorcerer's Stone, The Lord of the Rings: The Two Towers, Harry Potter and the Chamber of Secrets, Spider-Man, Finding Nemo, The Lord of the Rings: The Return of the King, The Matrix Reloaded, Harry Potter and the Prisoner of Azkaban, Shrek 2, Spider-Man 2, Star Wars: Episode III - Revenge of the Sith, Harry Potter and the Goblet of Fire, The Chronicles of Narnia: The Lion, the Witch and the Wardrobe, The Da Vinci Code, Pirates of the Caribbean: Dead Man's Chest, Transformers, Harry Potter and the Order of the Phoenix, Shrek the Third, Spider-Man 3, Pirates of the Caribbean: At World's End, Indiana Jones and the Kingdom of the Crystal Skull, The Dark Knight, Ice Age: Dawn of the Dinosaurs, Transformers: Revenge of the Fallen, Up, 2012, Harry Potter and the Half-Blood Prince, Inception, Shrek Forever After, Alice in Wonderland, Toy Story 3, Harry Potter and the Deathly Hallows: Part 1, The Twilight Saga: Breaking Dawn - Part 1, Harry Potter and the Deathly Hallows: Part 2, Transformers: Dark of the Moon \\
    \hline
    \end{tabular}
    \caption{Movies: Cluster 1 (Part 1)}
    \label{tab:movies_one_pone}
\end{table}

\begin{table}[ht]
    \centering
    \small
    \begin{tabular}{|p{1\textwidth}|}
    \hline
    \textbf{Cluster 1 Movies (Part 2)} \\
    \hline
    Pirates of the Caribbean: On Stranger Tides, Ice Age: Continental Drift, The Twilight Saga: Breaking Dawn - Part 2, Madagascar 3: Europe's Most Wanted, Skyfall, The Amazing Spider-Man, The Dark Knight Rises, The Hobbit: An Unexpected Journey, Despicable Me 2, Gravity, The Hunger Games: Catching Fire, Frozen, Fast \& Furious 6, Iron Man 3, Monsters University, The Hobbit: The Desolation of Smaug, The Hunger Games: Mockingjay - Part 1, Captain America: The Winter Soldier, Dawn of the Planet of the Apes, Guardians of the Galaxy, Maleficent, The Amazing Spider-Man 2, Transformers: Age of Extinction, X-Men: Days of Future Past, The Hobbit: The Battle of the Five Armies, Minions, Inside Out, Spectre, Avengers: Age of Ultron, Deadpool, The Secret Life of Pets, Zootopia, The Jungle Book, Suicide Squad, Fantastic Beasts and Where to Find Them, Finding Dory, Rogue One: A Star Wars Story, Batman v Superman: Dawn of Justice, Captain America: Civil War, Despicable Me 3, Wonder Woman, Jumanji: Welcome to the Jungle, Beauty and the Beast, Spider-Man: Homecoming, Coco, Thor: Ragnarok, Guardians of the Galaxy Vol. 2, Pirates of the Caribbean: Dead Men Tell No Tales, The Fate of the Furious, Star Wars: The Last Jedi, Bohemian Rhapsody, Deadpool 2, Venom, Aquaman, Jurassic World: Fallen Kingdom, Mission: Impossible - Fallout, Black Panther, Incredibles 2, Spider-Man: Far from Home, Captain Marvel, Aladdin, Toy Story 4, Fast \& Furious Presents: Hobbs \& Shaw, No Time to Die \\
    \hline
    \end{tabular}
    \caption{Movies: Cluster 1 (Part 2)}
    \label{tab:movies_one_ptwo}
\end{table}

\begin{table}[ht]
    \centering
    \begin{tabular}{|p{1\textwidth}|}
    \hline
    \textbf{Cluster 2 Movies (Part 1)} \\
    \hline
    Look Who's Talking, Driving Miss Daisy, Turner \& Hooch, Born on the Fourth of July, Field of Dreams, Uncle Buck, When Harry Met Sally..., Dead Poets Society, Parenthood, Lethal Weapon 2, The War of the Roses, National Lampoon' Christmas Vacation, Honey, I Shrunk the Kids, Ghostbusters II, The Little Mermaid, Back to the Future Part II, Tango \& Cash, Steel Magnolias, Three Men and a Little Lady, Teenage Mutant Ninja Turtles, Kindergarten Cop, Bird on a Wire, Misery, Edward Scissorhands, Presumed Innocent, Flatliners, The Hunt for Red October, Another 48 Hrs., Back to the Future Part III, Dick Tracy, The Godfather: Part III, Days of Thunder, Total Recall, Die Hard 2, Fried Green Tomatoes, The Silence of the Lambs, Sleeping with the Enemy, Father of the Bride, The Naked Gun 2 1/2: The Smell of Fear, Teenage Mutant Ninja Turtles II, Hot Shots!, City Slickers, Star Trek VI: The Undiscovered Country, The Addams Family, The Prince of Tides, What About Bob?, Cape Fear, JFK, Hook, Backdraft, The Crying Game, The Hand that Rocks the Cradle, Unforgiven, Wayne's World, White Men Can't Jump, Sister Act, Scent of a Woman, Lethal Weapon 3, Under Siege, Boomerang, A League of Their Own, The Last of the Mohicans, Bram Stoker's Dracula, A Few Good Men, Patriot Games, Batman Returns, Cool Runnings, Groundhog Day, Free Willy, Sleepless in Seattle, Schindler's List, Tombstone, Philadelphia, Dave, Grumpy Old Men, Indecent Proposal, Sister Act 2: Back in the Habit, In the Line of Fire, Rising Sun, The Firm, The Pelican Brief, Demolition Man, Cliffhanger, Pulp Fiction, Ace Ventura: Pet Detective, Dumb and Dumber, The Santa Clause, The Mask, Speed, Legends of the Fall, Star Trek: Generations, The Client, The Specialist, The Flintstones, Stargate, Disclosure, Interview with the Vampire, Clear and Present Danger, Wolf, Maverick, While You Were Sleeping, Dangerous Minds, Grumpier Old Men, Father of the Bride Part II, Ace Ventura: When Nature Calls, Get Shorty, Mr. Holland's Opus, Se7en, Casper, Congo, Crimson Tide, Pocahontas, Jumanji, Braveheart, Batman Forever, Waterworld, Scream, The First Wives Club, The English Patient, The Birdcage, Phenomenon, A Time to Kill, Star Trek: First Contact, Broken Arrow, Jerry Maguire, The Nutty Professor, 101 Dalmatians, The Rock, Ransom, Space Jam, The Hunchback of Notre Dame, Eraser, Michael, Good Will Hunting, I Know What You Did Last Summer, Scream 2, My Best Friend's Wedding, Liar Liar, As Good as It Gets, George of the Jungle, The Lost World: Jurassic Park  \\
    \hline
    \end{tabular}
    \caption{Movies: Cluster 2 (Part 1)}
    \label{tab:movies_two_pone}
\end{table}

\begin{table}[ht]
    \centering
    \begin{tabular}{|p{1\textwidth}|}
    \hline
    \textbf{Cluster 2 Movies (Part 2)} \\
    \hline
    Con Air, Conspiracy Theory, Face/Off, Flubber, Hercules, Air Force One, Contact, Tomorrow Never Dies, Batman and Robin, The Waterboy, Shakespeare in Love, The Rugrats Movie, Rush Hour, Stepmom, Patch Adams, The Truman Show, You've Got Mail, The Prince of Egypt, Doctor Dolittle, Deep Impact, Mulan, Enemy of the State, The Mask of Zorro, Lethal Weapon 4, The Blair Witch Project, American Pie, American Beauty, Big Daddy, The General's Daughter, The Green Mile, Runaway Bride, Double Jeopardy, Analyze This, Stuart Little, Wild Wild West, Crouching Tiger, Hidden Dragon, Scary Movie, Big Momma's House, Remember the Titans, Chicken Run, Traffic, Erin Brockovich, Meet the Parents, X-Men, Nutty Professor II: The Klumps, Charlie's Angels, What Lies Beneath, The Patriot, The Perfect Storm, American Pie 2, Spy Kids, The Princess Diaries, The Fast and the Furious, A Beautiful Mind, Vanilla Sky, Dr. Dolittle 2, Black Hawk Down, Lara Croft: Tomb Raider, Sweet Home Alabama, The Ring, Mr. Deeds, The Bourne Identity, Austin Powers in Goldmember, The Santa Clause 2, xxx, Lilo \& Stitch, Scooby-Doo, Bringing Down the House, Elf, Spy Kids 3-D: Game Over, Cheaper by the Dozen, Anger Management, 2 Fast 2 Furious, S.W.A.T., Something's Gotta Give, Seabiscuit, Bad Boys II, Hulk, Fahrenheit 9/11, DodgeBall: A True Underdog Story, The Village, 50 First Dates, The Bourne Supremacy, Lemony Snicket's A Series of Unfortunate Events, Van Helsing, The Polar Express, The 40 Year Old Virgin, Walk the Line, Wedding Crashers, Flightplan, The Pacifier, Robots, The Longest Yard, Fantastic Four, Fun with Dick and Jane, Chicken Little, Borat: Cultural Learnings of America for Make Benefit Glorious Nation of Kazakhstan, The Devil Wears Prada, Scary Movie 4, The Break-Up, The Pursuit of Happyness, Dreamgirls, Talladega Nights: The Ballad of Ricky Bobby, Over the Hedge, Click, The Departed, Juno, Knocked Up, Wild Hogs, Enchanted, American Gangster, Fantastic 4: Rise of the Silver Surfer, Rush Hour 3, Gran Torino, Marley and Me, Wanted, Get Smart, Horton Hears a Who!, The Incredible Hulk, Taken, Paul Blart: Mall Cop, The Blind Side, The Proposal, G.I. Joe: The Rise of Cobra, True Grit, Shutter Island, Grown Ups, Little Fockers, Megamind, The Last Airbender, The Help, Bridesmaids, Alvin and the Chipmunks: Chipwrecked, The Hangover Part II, Sherlock Holmes: A Game of Shadows, Lincoln, The Conjuring, Identity Thief, We're The Millers, American Hustle, The Heat, Neighbors, 22 Jump Street, Divergent, Straight Outta Compton, Pitch Perfect 2, The SpongeBob Movie: Sponge Out of Water, Hidden Figures, Ghostbusters, Get Out, The LEGO Batman Movie, A Quiet Place, Crazy Rich Asians, Glass, Us, The Upside, John Wick: Chapter 3 - Parabellum, The LEGO Movie 2: The Second Part, Tenet, The Father, A Quiet Place Part 2, Emma, Wonder Woman, Soul, Sonic the Hedgehog, Enola Holmes, Mulan, Greenland, West Side Story, Free Guy, Encanto, Tick, Tick... Boom!, Venom: Let There Be Carnage \\
    \hline
    \end{tabular}
    \caption{Movies: Cluster 2 (Part 2)}
    \label{tab:movies_two_ptwo}
\end{table}

\begin{table}[ht]
    \centering
    \begin{tabular}{|p{0.8\textwidth}|}
    \hline
    \textbf{Cluster 3 Movies} \\
    \hline
    Titanic, Avatar, The Avengers, Jurassic World, Furious 7, Star Wars: The Force Awakens, Avengers: Infinity War, The Lion King, Avengers: Endgame, Spiderman: No Way Home \\
    \hline
    \end{tabular}
    \caption{Movies: Cluster 3}
    \label{tab:movies_three}
\end{table}

\begin{table}[ht]
    \centering
    \begin{tabular}{|p{0.8\textwidth}|}
    \hline
    \textbf{Cluster 4 Movies (Part 1)} \\
    \hline
    Batman, Indiana Jones and the Last Crusade, Pretty Woman, Home Alone, Ghost, Dances with Wolves, Beauty and the Beast, Robin Hood: Prince of Thieves, Terminator 2: Judgment Day, Home Alone 2: Lost in New York, The Bodyguard, Aladdin, Basic Instinct, Mrs. Doubtfire, The Fugitive, Jurassic Park III, Forrest Gump, True Lies, Toy Story, Apollo 13, GoldenEye, Die Hard: With a Vengeance, Mission: Impossible, Twister, Men in Black, There's Something About Mary, Saving Private Ryan, A Bug's Life, Godzilla, Armageddon, The Sixth Sense, Notting Hill, The Matrix, The Mummy, Toy Story 2, The World Is Not Enough, Tarzan, What Women Want, Cast Away, Gladiator, How the Grinch Stole Christmas, Mission: Impossible II, Dinosaur, Shrek, Ocean's Eleven, Hannibal, Rush Hour 2, Jurassic Park III, The Mummy Returns, Planet of the Apes, Monsters, Inc., Pearl Harbor, My Big Fat Greek Wedding, Ice Age, Signs, Minority Report, Star Wars: Episode II - Attack of the Clones, Men in Black 2, Die Another Day, Bruce Almighty, X2: X-Men United, Pirates of the Caribbean: The Curse of the Black Pearl, The Last Samurai, The Matrix Revolutions, Terminator 3: Rise of the Machines, The Passion of the Christ, Shark Tale, Meet the Fockers, The Incredibles, National Treasure, Ocean's Twelve, I, Robot, The Day After Tomorrow, Troy, Hitch, Madagascar, Mr. \& Mrs. Smith, War of the Worlds, Batman Begins, Charlie and the Chocolate Factory, King Kong, Ice Age: The Meltdown, Happy Feet, Night at the Museum, Cars, Mission: Impossible III, Casino Royale, X-Men: The Last Stand, Superman Returns, 300, Alvin and the Chipmunks, The Bourne Ultimatum, The Simpsons Movie, Live Free or Die Hard, National Treasure: Book of Secrets, Ratatouille, I Am Legend, Slumdog Millionaire, Twilight, Mamma Mia!, Sex and the City, Kung Fu Panda, Iron Man, Hancock, Madagascar: Escape 2 Africa, The Curious Case of Benjamin Button, WALL-E, Quantum of Solace, The Chronicles of Narnia: Prince Caspian \\
    \hline
    \end{tabular}
    \caption{Movies: Cluster 4 (Part 1)}
    \label{tab:movies_four_pone}
\end{table}

\begin{table}[ht]
    \centering
    \begin{tabular}{|p{0.8\textwidth}|}
    \hline
    \textbf{Cluster 4: Movies (Part 2)} \\
    \hline
     The Hangover, The Twilight Saga: New Moon, Alvin and the Chipmunks: The Squeakquel, Fast and Furious, Sherlock Holmes, X-Men Origins: Wolverine, Star Trek, Night at the Museum: Battle of the Smithsonian, Monsters vs Aliens, The King's Speech, The Karate Kid, The Twilight Saga: Eclipse, Despicable Me, Clash of the Titans, How to Train Your Dragon, Tron Legacy, Iron Man 2, Tangled, Rio, Rise of the Planet of the Apes, The Smurfs, Fast Five, Puss in Boots: The Three Diablos, Captain America: The First Avenger, Mission: Impossible - Ghost Protocol, Thor, Kung Fu Panda 2, Cars 2, Taken 2, Ted, Les Miserables, The Lorax, The Hunger Games, Hotel Transylvania, Django Unchained, Wreck-It Ralph, Snow White and the Huntsman, Brave, Men in Black 3, The Great Gatsby, The Croods, Thor: The Dark World, Star Trek Into Darkness, Oz: The Great and Powerful, World War Z, Man of Steel, American Sniper, The Lego Movie, Gone Girl, Teenage Mutant Ninja Turtles, How to Train Your Dragon 2, Godzilla, Big Hero 6, Interstellar, Fifty Shades of Grey, Hotel Transylvania 2, Cinderella, The Martian, San Andreas, Ant-Man, Home, The Revenant, Mission: Impossible - Rogue Nation, The Hunger Games: Mockingjay - Part 2, La La Land, Sing, Jason Bourne, Trolls, Moana, Doctor Strange, X-Men: Apocalypse, Star Trek Beyond, It, The Greatest Showman, Logan, Dunkirk, The Boss Baby, Kong: Skull Island, Justice League, A Star is Born, Dr. Seuss' The Grinch, Hotel Transylvania 3: Summer Vacation, Spider-Man: Into the Spider-Verse, Mary Poppins Returns, Ralph Breaks the Internet, Ant-Man and the Wasp, Fantastic Beasts: The Crimes of Grindelwald, Solo: A Star Wars Story, It: Chapter Two, The Secret Life of Pets 2, Once Upon a Time in Hollywood, Shazam!, How to Train Your Dragon: The Hidden World, Pokemon Detective Pikachu, Dumbo, Godzilla: King of the Monsters, Hamilton, Dune \\
    \hline
    \end{tabular}
    \caption{Movies: Cluster 4 (Part 2)}
    \label{tab:movies_four_ptwo}
\end{table}

\begin{table}[h!]
    \centering
    \small 
    \setlength{\tabcolsep}{5pt} 
    \renewcommand{\arraystretch}{1} 
    \begin{tabular}{| m{1.2cm} | m{1.7cm} | m{1.7cm} | m{1.7cm} | m{1.7cm} |} 
        \hline
        \textbf{Cluster} & \textbf{t1-t5} & \textbf{t6-t10} & \textbf{t11-t15} & \textbf{t15-t19} \\ \hline
        1 & t1=32.039 t2=67.961 t3=7.767 t4=92.233 t5=0.0 & t6=100.0 t7=2.913 t8=27.184 t9=69.903 t10=5.825 & t11=44.66 t12=47.573 t13=1.942 t14=27.184 t15=72.816 & t16=35.922 t17=64.078 t18=13.592 t19=13.592 \\ \hline
        2 & t1=27.707 t2=72.293 t3=76.433 t4=23.567 t5=99.045 & t6=0.955 t7=56.051 t8=28.344 t9=15.605 t10=49.363 & t11=24.522 t12=16.879 t13=9.236 t14=61.146 t15=38.854 & t16=59.236 t17=40.764 t18=73.885 t19=73.885 \\ \hline
        3 & t1=10.0 t2=90.0 t3=0.0 t4=100.0 t5=0.0 & t6=100.0 t7=10.0 t8=10.0 t9=80.0 t10=20.0 & t11=70.0 t12=10.0 t13=0.0 t14=0.0 t15=100.0 & t16=10.0 t17=90.0 t18=10.0 t19=10.0 \\ \hline
        4 & t1=34.3 t2=65.7 t3=30.435 t4=69.565 t5=2.899 & t6=97.101 t7=17.391 t8=39.614 t9=42.995 t10=16.908 & t11=38.164 t12=39.13 t13=5.797 t14=45.894 t15=54.106 & t16=44.928 t17=55.072 t18=33.816 t19=33.816 \\ \hline
    \end{tabular}
    \caption{Movies: Tag Percentages}
    \label{tab:movies_tag_percentages}
\end{table}


\begin{table}[h!]
\centering
\small
\begin{adjustbox}{max width=\textwidth}
\begin{tabular}{|m{3cm}|m{12cm}|}
\hline
\textbf{Feature}         & \textbf{Values}                                                                                                                                       \\ \hline
\textbf{age}             & Continuous                                                                                                                                           \\ \hline
\textbf{workclass}       & Private, Self-emp-not-inc, Self-emp-inc, Federal-gov, Local-gov, State-gov, Without-pay, Never-worked                                                \\ \hline
\textbf{fnlwgt (Final Weight)}          & Continuous                                                                                                                                           \\ \hline
\textbf{education}       & Bachelors, Some-college, 11th, Hs-grad, Prof-school, Assoc-acdm, Assoc-voc, 9th, 7th-8th, 12th, Masters, 1st-4th, 16th, Doctorate, 5th-6th, Preschool \\ \hline
\textbf{education-num}   & Continuous                                                                                                                                           \\ \hline
\textbf{marital-status}  & Married-civ-spouse, Divorced, Never-married, Separated, Widowed, Married-spouse-absent, Married-AF-spouse                                             \\ \hline
\textbf{occupation}      & Tech-support, Craft-repair, Other-service, Sales, Exec-managerial, Prof-specialty, Handlers-cleaners, Machine-op-inspct, Adm-clerical, Farming-fishing, Transport-moving, Priv-house-serv, Protective-serv, Armed-Forces \\ \hline
\textbf{relationship}    & Wife, Own-child, Husband, Not-in-family, Other-relative, Unmarried                                                                                   \\ \hline
\textbf{race}            & White, Asian-Pac-Islander, Amer-Indian-Eskimo, Other, Black                                                                                          \\ \hline
\textbf{sex}             & Female, Male                                                                                                                                         \\ \hline
\textbf{capital-gain}    & Continuous                                                                                                                                           \\ \hline
\textbf{capital-loss}    & Continuous                                                                                                                                           \\ \hline
\textbf{hours-per-week}  & Continuous                                                                                                                                           \\ \hline
\textbf{native-country}  & United-States, Cambodia, England, Puerto-Rico, Canada, Germany, Outlying-US(Guam-USVI-etc), India, Japan, Greece, South, China, Cuba, Iran, Honduras, Philippines, Italy, Poland, Jamaica, Vietnam, Mexico, Portugal, Ireland, France, Dominican-Republic, Laos, Ecuador, Taiwan, Haiti, Columbia, Hungary, Guatemala, Nicaragua, Scotland, Thailand, Yugoslavia, El-Salvador, Trinidad \& Tobago, Peru, Hong, Holand-Netherlands \\ \hline
\textbf{income}          & >50K, <=50K                                                                                                                                          \\ \hline
\end{tabular}
\end{adjustbox}
\caption{Original Raw Features of Adult Census Dataset}
\label{tab:census_original_features}
\end{table}

\begin{table}[h!]
    \centering
    \small 
    \setlength{\tabcolsep}{5pt} 
    \renewcommand{\arraystretch}{1} 
    \begin{tabular}{| m{1.2cm} | m{1.7cm} | m{1.7cm} | m{1.7cm} |} 
        \hline
        \textbf{Cluster} & \textbf{t1-t7} & \textbf{t8-t14} & \textbf{t15-t21} \\ \hline
        1 & t1=0.173 t2=8.981 t3=90.846 t4=79.135 t5=20.865 t6=49.308 t7=50.692 & t8=28.788 t9=32.538 t10=38.673 t11=84.808 t12=3.096 t13=0.923 t14=10.308 & t15=0.865 t16=52.288 t17=47.712 t18=100.0 t19=0.0 t20=93.308 t21=6.692 \\ \hline
        2 & t1=0.035 t2=14.265 t3=85.7 t4=65.498 t5=34.502 t6=44.602 t7=55.398 & t8=28.883 t9=32.069 t10=39.048 t11=85.621 t12=3.155 t13=0.964 t14=9.445 & t15=0.815 t16=29.173 t17=70.827 t18=8.277 t19=91.723 t20=72.516 t21=27.484 \\ \hline
    \end{tabular}
    \caption{Census: Tag Percentages}
    \label{tab:census_tag_percentages}
\end{table}


\begin{table}[ht]
    \centering
    \small
    \begin{tabular}{|p{1\textwidth}|}
    \hline
    \textbf{Cluster 1 Fields of Study (Part 1)} \\
    \hline
    Construction Services, Commercial Art and Graphic Design, Hospitality Management, Cosmetology Services and Culinary Arts, Communication Technologies, Court Reporting, Agriculture Production and Management, Computer Programming and Data Processing, Advertising and Public Relations, Film Video and Photographic Arts, Electrical, Mechanical, and Precision Technologies and Production, Mechanical Engineering Related Technologies, Mass Media, Transportation Sciences and Technologies, Computer Networking and Telecommunications, Miscellaneous Business \& Medical Administration, Miscellaneous Engineering Technologies, Industrial Production Technologies, Miscellaneous Fine Arts, Criminology, Management Information Systems and Statistics, Computer Administration Management and Security, Operations Logistics and E-Commerce, Medical Technologies Technicians, Computer and Information Systems, Actuarial Science, Electrical Engineering Technology, Journalism, Medical Assisting Services, Engineering Technologies, Information Sciences, Architectural Engineering, Multi/Interdisciplinary Studies, Nuclear, Industrial Radiology, and Biological Technologies, General Agriculture, Forestry, Human Services and Community Organization, Visual and Performing Arts, Natural Resources Management, Studio Arts, Family and Consumer Sciences, Physical Fitness Parks Recreation and Leisure, Petroleum Engineering, Plant Science and Agronomy, Human Resources and Personnel Management, International Business, Composition and Rhetoric, Drama and Theater Arts, Business Economics, Engineering and Industrial Management, Health and Medical Administrative Services, Agricultural Economics, Environmental Science, Geography, Miscellaneous Engineering, Ecology, Interdisciplinary Social Sciences, Architecture, Soil Science, Pre-Law and Legal Studies, Civil Engineering, Computer Engineering, Mining and Mineral Engineering, Early Childhood Education, General Social Sciences, Animal Sciences, Treatment Therapy Professions, Miscellaneous Agriculture, Humanities, Food Science, Industrial and Manufacturing Engineering, Geological and Geophysical Engineering \\
    \hline
    \end{tabular}
    \caption{College Majors: Cluster 1 (Part 1)}
    \label{tab:majors_one_pone}
\end{table}

\begin{table}[ht]
    \centering
    \small
    \begin{tabular}{|p{1\textwidth}|}
    \hline
    \textbf{Cluster 1 Fields of Study (Part 2)} \\
    \hline
    Social Psychology, Naval Architecture and Marine Engineering, Mathematics and Computer Science, Art History and Criticism, Miscellaneous Health Medical Professions, General Medical and Health Services, Intercultural and International Studies, Nutrition Sciences, Physical and Health Education Teaching, Community and Public Health, Theology and Religious Vocations, Oceanography, Miscellaneous Education, Biological Engineering, Public Administration, Industrial and Organizational Psychology, Military Technologies, Music, Art and Music Education, Linguistics and Comparative Language and Literature, Materials Engineering and Materials Science, Anthropology and Archeology, Social Work, Teacher Education: Multiple Levels, Geology and Earth Science, Pharmacy Pharmaceutical Sciences and Administration, Other Foreign Languages, Area Ethnic and Civilization Studies, Physical Sciences, Atmospheric Sciences and Meteorology, Chemical Engineering, Aerospace Engineering, Miscellaneous Social Sciences, Applied Mathematics, Statistics and Decision Science, French German Latin and Other Common Foreign Language Studies, Social Science or History Teacher Education, International Relations, Environmental Engineering, Miscellaneous Biology, Miscellaneous Psychology, Metallurgical Engineering, Secondary Teacher Education, Geosciences, United States History, Engineering Mechanics Physics and Science, Cognitive Science and Biopsychology, Language and Drama Education, Nuclear Engineering, Public Policy, Mathematics Teacher Education, Science and Computer Teacher Education, Microbiology, Philosophy and Religious Studies, Special Needs Education, Botany, Astronomy and Astrophysics, Physiology, Biomedical Engineering, Library Science, Molecular Biology, Pharmacology, Zoology, Physics, Neuroscience, Educational Psychology, Biochemical Sciences, Genetics, Materials Science, Communication Disorders Sciences and Services, Counseling Psychology, Clinical Psychology, Health and Medical Preparatory Programs, School Student Counseling, Educational Administration and Supervision \\
    \hline
    \end{tabular}
    \caption{College Majors: Cluster 1 (Part 2)}
    \label{tab:majors_one_ptwo}
\end{table}

\begin{table}[ht]
    \centering
    \small
    \begin{tabular}{|p{1\textwidth}|}
    \hline
    \textbf{Cluster 2 Fields of Study} \\
    \hline
    Marketing and Marketing Research, Criminal Justice and Fire Protection, Communications, Fine Arts, Liberal Arts, Finance, Computer Science, General Engineering, Multi-Disciplinary or General Science, Sociology, Mechanical Engineering, Economics, Electrical Engineering, English Language and Literature, History, Mathematics, Political Science and Government, Biology, Chemistry \\
    \hline
    \end{tabular}
    \caption{College Majors: Cluster 2}
    \label{tab:majors_two}
\end{table}

\begin{table}[ht]
    \centering
    \small
    \begin{tabular}{|p{1\textwidth}|}
    \hline
    \textbf{Cluster 3 Fields of Study} \\
    \hline
    Business Management and Administration, General Business, Accounting, Nursing, Elementary Education, General Education, Psychology \\
    \hline
    \end{tabular}
    \caption{College Majors: Cluster 3}
    \label{tab:majors_three}
\end{table}

\begin{table}[h!]
    \centering
    \small 
    \setlength{\tabcolsep}{5pt} 
    \renewcommand{\arraystretch}{1} 
    \begin{tabular}{| m{1.2cm} | m{1.7cm} | m{1.7cm} | m{1.7cm} | m{1.7cm} | m{1.7cm} | m{1.7cm} |}
        \hline
        \textbf{Cluster} & \textbf{t1-t5} & \textbf{t6-t10} & \textbf{t11-t15} & \textbf{t16-t20} & \textbf{t21-t25} & \textbf{t26-t30} \\ \hline
        1 & t1=76.19 t2=23.81 t3=58.503 t4=41.497 t5=58.503 & t6=41.497 t7=58.503 t8=41.497 t9=58.503 t10=41.497 & t11=58.503 t12=41.497 t13=46.939 t14=53.061 t15=52.381 & t16=47.619 t17=58.503 t18=41.497 t19=58.503 t20=41.497 & t21=58.503 t22=41.497 t23=58.503 t24=41.497 t25=49.66 & t26=50.34 t27=48.98 t28=51.02 t29=50.34 t30=49.66 \\ \hline
        2 & t1=73.684 t2=26.316 t3=0.0 t4=100.0 t5=0.0 & t6=100.0 t7=0.0 t8=100.0 t9=0.0 t10=100.0 & t11=0.0 t12=100.0 t13=31.579 t14=68.421 t15=31.579 & t16=68.421 t17=0.0 t18=100.0 t19=0.0 t20=100.0 & t21=0.0 t22=100.0 t23=0.0 t24=100.0 t25=42.105 & t26=57.895 t27=26.316 t28=73.684 t29=42.105 t30=57.895 \\ \hline
        3 & t1=100.0 t2=0.0 t3=0.0 t4=100.0 t5=0.0 & t6=100.0 t7=0.0 t8=100.0 t9=0.0 t10=100.0 & t11=0.0 t12=100.0 t13=42.857 t14=57.143 t15=42.857 & t16=57.143 t17=0.0 t18=100.0 t19=0.0 t20=100.0 & t21=0.0 t22=100.0 t23=0.0 t24=100.0 t25=42.857 & t26=57.143 t27=42.857 t28=57.143 \\ \hline
    \end{tabular}
    \caption{College Majors: Tag Percentages}
    \label{tab:majors_tag_percentages}
\end{table}


\end{document}